\documentclass[preprint,authoryear,12pt]{elsarticle}

\usepackage[T1]{fontenc}
\usepackage{xcolor}
\usepackage{url}
\usepackage{booktabs}
\usepackage{threeparttable} % width-matched table notes in the appendix
\usepackage{amsmath,amssymb}
\usepackage{bm}
\usepackage{multirow}
\usepackage{hyperref}
\usepackage{orcidlink}       % \orcidlink{} ORCID iD icon next to author names
\usepackage{float}           % [H] placement for floats
\usepackage{placeins}        % \FloatBarrier to keep floats in their section
\usepackage{dblfloatfix}     % allow table*/figure* on their defining page in two-column mode
\usepackage{tikz}
\usetikzlibrary{arrows.meta,positioning,shapes.geometric,calc,decorations.pathreplacing}

\begin{document}
\begin{frontmatter}

%%, - title,-

\title{Gradient Concentration, Not Weight Saliency, Explains Representation-Level Class Unlearning}

%%, - authors,-
\author[dibris]{Billel~Habbati\orcidlink{0009-0000-8315-7182}\fnref{fn-dibris}}
\ead{billel.habbati@edu.unige.it}

\author[casd]{Alessio~Merlo\orcidlink{0000-0002-2272-2376}\corref{cor1}\fnref{fn-casd}}
\ead{alessio.merlo@unicasd.it}

\author[dibris]{Luca~Verderame\orcidlink{0000-0001-7155-7429}\fnref{fn-dibris}}
\ead{luca.verderame@unige.it}

\author[dibris]{Meriem~Guerar\orcidlink{0000-0003-4566-1382}\fnref{fn-dibris}}
\ead{meriem.guerar@unige.it}

%%, - corresponding author,-
\cortext[cor1]{Corresponding author.}

%%, - author footnotes / thanks,-
\fntext[fn-dibris]{B.~Habbati, L.~Verderame, and M.~Guerar are with DIBRIS --
  Department of Informatics, Bioengineering, Robotics and Systems Engineering,
  University of Genova, Via Dodecaneso 35, 16146 Genova, Italy.
  B.~Habbati is a Ph.D.\ student.
  E-mail: billel.habbati@edu.unige.it, luca.verderame@unige.it,
  meriem.guerar@unige.it.}

\fntext[fn-casd]{A.~Merlo is with the Centre for Defense Higher Studies (CASD),
  Palazzo Salviati, Piazza della Rovere 83, 00165 Roma, Italy.
  E-mail: alessio.merlo@unicasd.it. Corresponding author: A.~Merlo.}

\address[dibris]{DIBRIS -- Department of Informatics, Bioengineering, Robotics and Systems Engineering, University of Genova, Via Dodecaneso 35, 16146 Genova, Italy}
\address[casd]{Centre for Defense Higher Studies (CASD), Palazzo Salviati, Piazza della Rovere 83, 00165 Roma, Italy}

%%, - abstract,-

\begin{abstract}
Machine unlearning aims to remove the influence of specific training data while preserving model utility. Many state-of-the-art approaches pursue this goal by restricting the forgetting update to a subset of parameters selected through gradient-based saliency. Although such methods are widely adopted, the actual contribution of saliency-based weight selection to representation-level forgetting remains unclear. In this work, we perform the first controlled ablation of the saliency masking mechanism used by SalUn. Using a matched-compute experimental design on CIFAR-10 and CIFAR-100 with ResNet-18, we compare saliency-based masking against random masks of equal sparsity and unconstrained updates, while keeping the unlearning objective, optimization schedule, and computational budget fixed. Across multiple representation-level evaluations, including linear probing, prototype recovery, and layer-wise CKA, the three configurations exhibit statistically equivalent representation-level recoverability. We find that forget gradients are strongly concentrated in the final network layers ($\approx92\%$ of the squared gradient energy on CIFAR-10) before any mask is applied, causing all masking strategies to operate within the same representational subspace. Furthermore, saliency masks show limited class specificity (specificity index $0.09$--$0.11$), selecting highly overlapping parameter subsets across different forget classes. Our findings suggest that, in the studied setting, representation-level forgetting is primarily governed by gradient concentration and representation geometry rather than by the specific identity of saliency-selected weights. More broadly, the results support a growing body of evidence indicating that effective representation-level unlearning requires objectives that act directly on latent representations rather than on increasingly sophisticated weight-selection strategies.

\end{abstract}

%%, - keywords,-


%%, - highlights (3-5 bullets, <=85 chars each; indexed on ScienceDirect),-

\begin{highlights}
\setlength{\itemsep}{0pt}
\setlength{\parskip}{0pt}
\item A matched-compute ablation isolates the role of saliency-based masking.
\item Mask identity does not explain representation-level class recoverability.
\item Forget gradients carry $\approx92\%$ of their energy in late layers before masking.
\item Saliency masks exhibit weak class specificity (specificity index $0.09$--$0.11$).
\item Effective unlearning may require objectives acting directly on representations.
\vspace{\fill}
\end{highlights}

\end{frontmatter}

%%, - (line numbers disabled for the public / arXiv preprint),-

%%======================================================================
%% MAIN TEXT
%%======================================================================

\section{Introduction}

\label{sec:intro} 

Machine unlearning seeks to remove the influence of specific training data from a learned model without requiring retraining from scratch. The problem has attracted significant attention because modern AI systems are increasingly expected to support data deletion requests, comply with privacy regulations such as the GDPR, and adapt to evolving legal and operational constraints. While exact retraining remains the gold standard for data removal, its computational cost is often prohibitive for large-scale models and frequent deletion requests. As a result, most practical approaches rely on approximate unlearning strategies that modify an already-trained model in place while attempting to preserve utility on the retained data \citep{bourtoule2021sisa,voigt2017gdpr,xu2024survey}. 

A growing body of evidence, however, suggests that evaluating unlearning solely through output behavior can be misleading. Several recent studies have shown that a model may achieve near-zero accuracy on forgotten data while still retaining substantial information about that data within its internal representations \citep{gao2026illusion,kim2025forgetting,ha2025blindspots}. In such cases, supposedly forgotten concepts can often be recovered through simple attacks such as linear probing or feature-space reconstruction. These findings have shifted attention from output-level forgetting toward representation-level forgetting, raising a broader question: what mechanisms actually drive forgetting inside modern unlearning methods?

One influential family of methods addresses forgetting through selective parameter updates. Rather than modifying all model weights during unlearning, these approaches first identify a subset of parameters relevant to the information to be forgotten and restrict optimization to that subset. SalUn \citep{fan2024salun}, a representative example of this paradigm, computes a gradient-based saliency mask and updates only the weights selected by that mask. The underlying intuition is appealing: if forgetting-relevant parameters can be identified accurately, modifying only those parameters should improve both efficiency and deletion quality. 

Despite the widespread adoption of saliency-based masking, its actual role remains surprisingly unclear. Existing studies typically compare SalUn against alternative unlearning methods, but rarely isolate the masking mechanism itself. Consequently, the effect of weight selection remains confounded with the forgetting objective that operates on top of it. This leaves open a fundamental scientific question: 
\emph{Does saliency-based weight selection truly drive representation-level forgetting, or would any mask of comparable size produce the same outcome?} 

In this paper, we answer this question through a controlled matched-compute ablation. Starting from the same pretrained model, we evaluate three conditions that differ only in the identity of the updated parameters: (i) SalUn's gradient-saliency mask, (ii) a random mask with identical sparsity, and (iii) unconstrained fine-tuning without masking. All other factors, including the unlearning objective, optimization schedule, and computational budget, are held constant. This design allows us to isolate the contribution of mask selection itself.

\begin{figure}[t]
\centering
\includegraphics[width=0.98\linewidth]{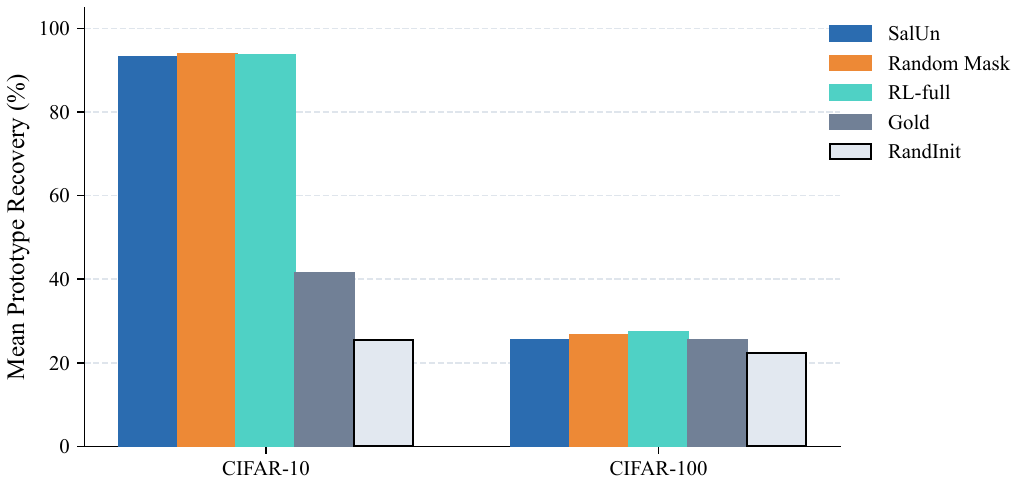}
\caption{\textbf{The saliency mask does not change what survives.}
Mean forget-class prototype recovery ($k{=}5$) across ten classes.
On CIFAR-10 all three conditions are equivalent (TOST, Table~\ref{tab:tost}) and sit ${\approx}52$pp above Gold;
on CIFAR-100 they cluster near Gold, which itself does little to erase fine-grained classes.
\textit{RandInit}: untrained network (texture floor). Higher~$=$~more recoverable.}
\label{fig:teaser}
\end{figure}

Our analysis reveals three main findings. First, saliency-based masking, random masking, and unconstrained updates produce statistically equivalent representation-level recoverability on both CIFAR-10 and CIFAR-100 (Figure~\ref{fig:teaser}). Second, this equivalence is explained by the structure of the forgetting gradients: before any mask is applied, most gradient energy is already concentrated in the final layers of the network, causing all masking strategies to operate within essentially the same representational subspace. Third, saliency masks exhibit limited class specificity, selecting highly overlapping parameter subsets across different forget classes. Taken together, these results suggest that representation-level forgetting is governed primarily by gradient concentration and representation geometry rather than by the precise identity of saliency-selected weights. 

Beyond the specific case of SalUn, our findings contribute to a broader discussion emerging in recent representation-level unlearning research. Contemporary methods increasingly achieve forgetting by manipulating latent representations directly rather than by refining parameter selection strategies. Our results provide a mechanistic explanation for this trend by showing that weight selection alone may be insufficient to alter the underlying geometry that encodes forgotten concepts. 

Unlike the large and growing body of work that proposes new unlearning \emph{methods}, this paper provides a controlled mechanistic \emph{analysis} of an existing and widely used one. To the best of our knowledge, this is the first study to isolate SalUn-style saliency masking under a matched-compute protocol and to trace its representation-level behavior to gradient concentration and the limited class specificity of the selected masks. Our aim is not to introduce another deletion algorithm, but to explain why a popular mechanism behaves as it does, a question with direct consequences for how future representation-level unlearning methods are designed and evaluated.

The main contributions of this paper are:

\begin{itemize} 
\item We present the first controlled ablation of saliency-based masking in class unlearning, isolating the contribution of weight selection from that of the underlying unlearning objective.

\item We show that saliency masks, random masks of equal sparsity, and unconstrained updates yield statistically equivalent representation-level recoverability outcomes across multiple evaluation protocols.

\item We identify gradient concentration in late network layers as the primary mechanism explaining this equivalence: it is not tied to one selection rule (a diagonal-Fisher mask reproduces the result and selects nearly the same weights, Jaccard $0.83$--$0.94$), and it is consistent with the saliency mask's weak class specificity (index $0.09$--$0.11$) and heavy overlap across different forgotten classes.

\item We run a causal placement intervention that forcibly excludes the late subspace and find that output-level forgetting does not require it, while linear separability of the forgotten class survives under every placement; late-subspace updates control only prototype geometry, not erasure. These findings bear directly on the design of future unlearning methods and on the evaluation of representation-level deletion guarantees.

\end{itemize}

\section{Related Work}

\label{sec:related}

\subsection{Machine Unlearning Beyond Output-Level Forgetting}

Machine unlearning aims to remove the influence of specific training data from a trained model while avoiding the cost of full retraining. The problem was first framed for statistical-query learners by Cao and Yang \citep{caoyang2015}. Early work focused primarily on exact or certified removal mechanisms, including data sharding, efficient retraining procedures, influence-function-based approaches, and deletion guarantees for convex and smooth models \citep{bourtoule2021sisa,ginart2019deletion,guo2020certified,koh2017influence,sekhari2021remember,neel2021descent,izzo2021approximate}.
Because these approaches remain difficult to apply to large deep neural networks, much of the recent literature has shifted toward approximate unlearning methods that modify an existing model directly \citep{nguyen2022survey,xu2024survey}, with the large-language-model setting surveyed separately \citep{li2025llmunlearning}. 

The effectiveness of approximate unlearning has traditionally been evaluated through output-level metrics, such as forget-set accuracy and membership inference attacks \citep{shokri2017membership,yeom2018privacy,carlini2022lira}. However, recent studies have questioned whether these metrics adequately capture information removal. Several studies have shown that models can achieve apparent forgetting while still retaining substantial information in their internal representations \citep{hayes2024evaluations,gao2026illusion,kim2025forgetting,ha2025blindspots}. Beyond representation probing, apparent forgetting has also been shown to fail against data poisoning \citep{pawelczyk2024poisoning} and to be reversible from a few corrective samples \citep{goel2024corrective}, motivating more careful unlearning benchmarks \citep{grimes2024forgotten,thudi2022unrolling}. This observation has motivated growing interest in representation-level evaluation as a complementary perspective on unlearning.

\subsection{Parameter Selection and Saliency-Based Unlearning}

A large family of approximate unlearning methods performs targeted modifications of model parameters rather than retraining the entire model. Existing approaches include gradient-based forgetting, amnesiac relabeling, decision-boundary modification, teacher--student distillation, selective synaptic dampening, error-maximizing noise, logit-based filtration, and feature-oriented forgetting strategies \citep{golatkar2020eternal,graves2021amnesiac,chen2023boundary,kurmanji2023scrub,chundawat2023badteacher,foster2024ssd,tarun2023unsir,baumhauer2022linear,warnecke2023features}. 

Within this family, SalUn \citep{fan2024salun} introduced gradient-based weight saliency to identify parameters presumed most relevant to the information being forgotten. The central intuition is that restricting updates to a carefully selected subset of weights can improve the efficiency of forgetting while preserving model utility. Related ideas have also appeared in sparsity-aware learning and pruning-inspired approaches, where only a subset of parameters is considered critical for adaptation or retention \citep{jia2023sparse,frankle2019lottery,han2015learning,mallya2018packnet}. 

Despite their popularity, saliency-based strategies are typically evaluated within complete unlearning pipelines. As a consequence, the individual contribution of parameter selection remains difficult to separate from that of the optimization objective used during forgetting.

\subsection{Representation-Level Unlearning}

Recent work has increasingly focused on removing information directly from learned representations rather than only modifying model outputs. Gao et al. \citep{gao2026illusion} demonstrated that several state-of-the-art unlearning methods leave class information highly recoverable through linear probing, highlighting a gap between output-level forgetting and representation-level erasure. Similar observations have been reported by subsequent analyses of feature-space recoverability and representation memorization \citep{kim2025forgetting,yong2026erased,almudevar2026representation}, and information-theoretic metrics have been proposed to quantify erasure strength \citep{jeon2026idi}. 

Motivated by these findings, newer approaches have begun to optimize forgetting directly in latent space. Examples include projection-based methods derived from neural-collapse geometry \citep{papyan2020neural,pour2025representation}, contrastive representation shaping \citep{tang2026clreg}, and intermediate-layer representation editing \citep{lee2026erase}. Other work has explored depth-aware identification and removal of forget-specific directions across network layers \citep{hatami2026damp}. Representational similarity between networks is quantified with tools such as linear CKA \citep{kornblith2019cka}, SVCCA \citep{raghu2017svcca}, and canonical-correlation variants \citep{morcos2018insights}, which we use to localize where unlearning reshapes the representation. Collectively, these studies suggest that representation-level forgetting may depend more strongly on feature geometry than on output behavior alone.

\subsection{Research Gap}

Although prior work has extensively compared alternative unlearning algorithms and increasingly investigated representation-level evaluation, one important question remains unanswered: \emph{What is the actual contribution of saliency-based parameter selection to representation-level forgetting?} 

Existing studies evaluate saliency masks as components of complete unlearning methods, making it difficult to disentangle the effect of weight selection from that of the forgetting objective. Consequently, it remains unclear whether representation-level forgetting is primarily driven by the choice of masked parameters, by the optimization objective itself, or by structural properties of the underlying gradient dynamics. This paper addresses that gap through a controlled component-level ablation that isolates the role of saliency-based masking while keeping all other aspects of the unlearning process fixed.

\section{Methodology}

\label{sec:method} 

This section describes the experimental design used to isolate the role of saliency-based weight selection in class unlearning. We first introduce the unlearning setting and notation, then define the three matched-compute conditions used in the ablation, and finally describe the representation-level metrics and statistical protocol.

\subsection{Problem setting and notation}

Let \(f_{\bm{\theta}}\) be a classifier with parameters \(\bm{\theta}\in\mathbb{R}^{p}\), trained on a dataset \[ \mathcal{D}=\mathcal{D}_r\cup\mathcal{D}_f , \] 
where \(\mathcal{D}_f\) is the forget set and \(\mathcal{D}_r\) is the retain set. 
In this work, \(\mathcal{D}_f\) contains all samples from one target class, while \(\mathcal{D}_r\) contains the remaining classes. 

The gold-standard unlearned model is obtained by retraining from scratch only on the retain set:

\begin{equation} 
	\bm{\theta}^{\star} = \arg\min_{\bm{\theta}} \mathcal{L}_{r}(\bm{\theta}), \qquad \mathcal{L}_{r}(\bm{\theta}) = \frac{1}{|\mathcal{D}_r|} \sum_{(x,y)\in\mathcal{D}_r} \ell\!\left(f_{\bm{\theta}}(x),y\right).
\label{eq:gold}
\end{equation} 
Approximate unlearning seeks a parameter vector \(\bm{\theta}_u\) that behaves like \(\bm{\theta}^{\star}\) with respect to the forgotten class, while preserving predictive utility on \(\mathcal{D}_r\). Since full retraining is expensive, practical methods instead modify the original trained model in place.

\subsection{Random-label unlearning objective}

We focus on the random-label (RL) objective \citep{golatkar2020eternal} used in SalUn-style class unlearning. The objective combines a forget term, which pushes samples in \(\mathcal{D}_f\) away from their original class, with a retain term, which preserves performance on \(\mathcal{D}_r\):

\begin{equation} 
	\mathcal{L}_{\mathrm{RL}}(\bm{\theta}) = \frac{1}{|\mathcal{D}_f|} \sum_{(x,y)\in\mathcal{D}_f} \ell\!\left(f_{\bm{\theta}}(x),\tilde{y}\right) + \lambda \frac{1}{|\mathcal{D}_r|} \sum_{(x,y)\in\mathcal{D}_r} \ell\!\left(f_{\bm{\theta}}(x),y\right),
\label{eq:rl}
\end{equation} 

where \(\tilde{y}\) is a randomly sampled incorrect label, i.e., \[ \tilde{y}\sim \mathrm{Unif}\bigl(\mathcal{Y}\setminus\{y\}\bigr), \] and \(\lambda>0\) controls the retain--forget trade-off. 

The first term reduces the model's association between forgotten samples and their original class, while the second term discourages unnecessary degradation of retained classes. Importantly, this objective is kept fixed in all our experimental conditions. Therefore, any difference between conditions can be attributed to the parameter mask rather than to the loss being optimized.

\subsection{Saliency mask and masked updates}

SalUn restricts the update to a subset of parameters selected by a gradient-based saliency mask. To define this mask rigorously, let \[ g_f(\bm{\theta}) = \nabla_{\bm{\theta}}\mathcal{L}_{f}(\bm{\theta}) \in\mathbb{R}^{p} \] be the gradient of the forget loss, where

\begin{equation} 
	\mathcal{L}_{f}(\bm{\theta}) = \frac{1}{|\mathcal{D}_f|} \sum_{(x,y)\in\mathcal{D}_f} \ell\!\left(f_{\bm{\theta}}(x),y\right).
\label{eq:forget_loss}
\end{equation} 

For a target sparsity level \(s\in(0,1]\), the saliency mask selects the \(k=\lfloor sp\rfloor\) parameters with largest absolute forget-gradient magnitude:

\begin{equation} 
m_{S,i} = \mathbf{1} \left[ |g_{f,i}(\bm{\theta})| \ge \tau_s\!\left(|g_f(\bm{\theta})|\right) \right], \qquad i=1,\ldots,p,
\label{eq:saliency_mask}
\end{equation} 
where \(\tau_s(|g_f(\bm{\theta})|)\) denotes the threshold selecting the top \(k=\lfloor sp\rfloor\) entries of the coordinate-wise absolute gradient vector. 
Equivalently, 
\[ \|\mathbf{m}_S\|_0 = k . \] 

Given a binary mask \(\mathbf{m}\in\{0,1\}^{p}\), the masked unlearning update is
\begin{equation} \bm{\theta}_{t+1} = \bm{\theta}_{t} - \eta \left( \mathbf{m} \odot \nabla_{\bm{\theta}} \mathcal{L}_{\mathrm{RL}}(\bm{\theta}_{t}) \right),
\label{eq:mask}
\end{equation}

where \(\eta\) is the learning rate and \(\odot\) denotes element-wise multiplication. 

This formulation makes explicit that the mask does not change the loss function. It only determines which gradient coordinates are allowed to update the model parameters.

\subsection{Matched-compute ablation}

\begin{figure*}[t]
\centering
\includegraphics[width=0.85\textwidth]{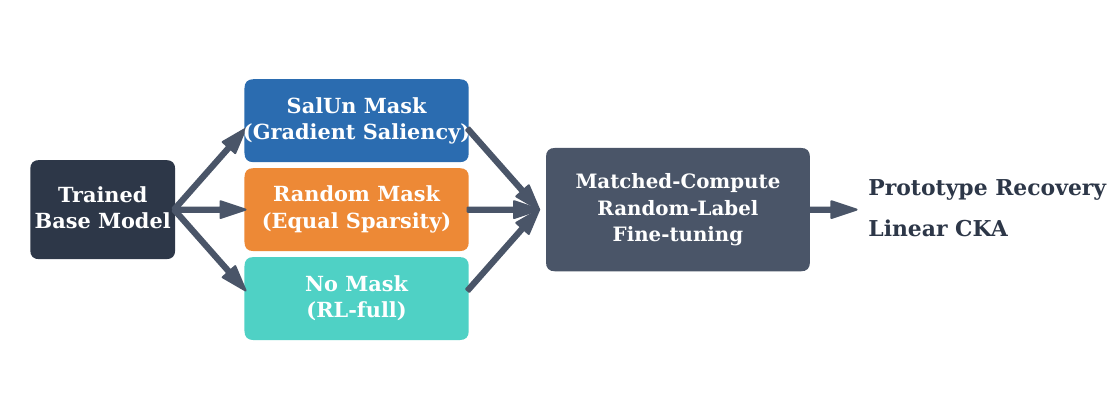}
\caption{\textbf{The matched-compute ablation.}
Three conditions branch from one base model and differ only in the weight mask;
learning rate, epochs, and retain term are identical.
Any difference in outcome is therefore attributable to mask selection alone.}
\label{fig:ablation}
\end{figure*}

The central goal of our study is to determine whether the identity of the saliency-selected weights is responsible for representation-level forgetting. To isolate this factor, we compare three conditions that differ only in the mask \(\mathbf{m}\) used in Eq.~\ref{eq:mask}:

\begin{itemize} 
	\item \textbf{Saliency mask}: \(\mathbf{m}=\mathbf{m}_S\), where \(\mathbf{m}_S\) selects the parameters with largest forget-gradient magnitude, as in Eq.~\ref{eq:saliency_mask}.

\item \textbf{Random mask}: \(\mathbf{m}=\mathbf{m}_R\), where \(\mathbf{m}_R\) is sampled uniformly among all binary masks with the same cardinality as the saliency mask: \[ \|\mathbf{m}_R\|_0=\|\mathbf{m}_S\|_0 . \] 
	\item \textbf{RL-full}: \(\mathbf{m}=\mathbf{1}\), corresponding to random-label unlearning without masking.

\end{itemize} All three conditions start from the same trained model and optimize the same objective in Eq.~\ref{eq:rl}. They also use the same learning rate, number of repair epochs, retain regularization, and data splits. Thus, the only controlled variable is the choice of mask (Figure~\ref{fig:ablation}). This design separates the effect of saliency-based parameter selection from the effect of the random-label unlearning objective. If the saliency mask is the main driver of representation-level forgetting, replacing it with a random mask or removing it should produce measurably different representation-level outcomes. If the three conditions behave similarly, then the results indicate that mask identity is not the dominant factor under this experimental setting.

Two further conditions run under the identical protocol and sharpen the analysis.
The \textbf{Fisher mask} replaces the saliency criterion with a second selection rule,
the top-$50\%$ of weights by diagonal empirical Fisher importance
(the second-moment statistic used by the SSD/EWC family \citep{foster2024ssd,kirkpatrick2017overcoming}),
so the claim is not tied to one mask construction.
The \textbf{Early mask} is a causal intervention on mask \emph{placement}: every weight in
the stem and the first three residual blocks is trainable and layer~4 plus the head are
frozen exactly as masked-out weights are above, forcing the update into the subspace
that no gradient-derived mask selects (\S\ref{sec:causal}); a forced variant repeats it
at $3\times$ the learning rate under gradient clipping.

\subsection{Datasets, model, and reference conditions}
We evaluate single-class unlearning on CIFAR-10 and CIFAR-100 \citep{krizhevsky2009cifar} using ResNet-18 \citep{he2016resnet}, the benchmark used by the closest representation-level auditing work \citep{gao2026illusion,jung2025opc,ha2025blindspots}. For each dataset, one class is treated as \(\mathcal{D}_f\), and all remaining classes form \(\mathcal{D}_r\). We use the same backbone, data splits, learning rate, repair epochs, and sparsity level across the matched conditions. In addition to the three unlearning conditions, we use two reference models:

\begin{itemize} \item \textbf{Gold}: a model retrained from scratch on \(\mathcal{D}_r\) (Eq.~\ref{eq:gold}), corresponding to exact class removal.

\item \textbf{RandInit}: an untrained network used as a lower-level feature baseline. This reference helps identify cases where prototype recovery reflects low-level visual or texture structure rather than retained class-specific knowledge.

\end{itemize} These references are not part of the matched-compute ablation. Instead, they provide context for interpreting how much information remains recoverable after unlearning.

\subsection{Representation-level evaluation}

We evaluate unlearning at both the output and representation levels. Output-level forget accuracy verifies whether the model has stopped predicting the forgotten class. However, because low forget accuracy does not necessarily imply removal of internal information, our main analysis focuses on representation-level recoverability. We use the following metrics:

\begin{itemize} \item \textbf{Linear probe}: a logistic regression classifier trained on frozen average-pooling features. This measures whether class information remains linearly accessible in the representation.

\item \textbf{Prototype recovery} \citep{ha2025blindspots}: a few-shot nearest-centroid attack in the penultimate feature space. A centroid is built from \(k\) forget-class examples, and held-out forget examples are classified by proximity to that centroid.

\item \textbf{Layer-wise CKA} \citep{kornblith2019cka}: linear centered kernel alignment between the original model and the unlearned model at each network depth. This measures which layers are most affected by unlearning.

\item \textbf{Head-only relearning}: a weaker attack that retrains only the final classifier head on a held-out retain subset (with no forget labels, so $0\%$ recall is expected) and measures whether the forgotten class reappears.

\item \textbf{Membership inference} \citep{shokri2017membership}: a permutation-test output-level sanity check based on standard membership-inference evaluation.

\end{itemize} Together, these metrics distinguish output-level forgetting from representation-level erasure. In particular, the linear probe and prototype recovery directly test whether the forgotten class remains recoverable from internal features.

\subsection{Statistical protocol}
The main empirical claim tested in this paper is one of equivalence rather than difference. A standard difference test can fail to reject the null hypothesis even when two methods are not meaningfully equivalent. Therefore, we use the two one-sided tests (TOST) procedure \citep{schuirmann1987tost,lakens2017equivalence} to assess whether the differences between conditions fall within a pre-specified equivalence margin. TOST appears not to have been used for equivalence claims in the unlearning literature, where ``similar'' is typically asserted by inspection. For each paired comparison, we test equivalence at margin \[ \delta = \pm 5 \text{pp}, \] where pp denotes percentage points. We fix $\delta$ \emph{a priori}, calibrated to the seed-noise floor: it sits at the level of the inter-seed standard deviation of recovery ($\approx3$--$4$pp on CIFAR-10), so a difference inside the margin cannot be distinguished from run-to-run noise, while it remains below any difference meaningful to an attacker. A significant TOST result indicates that the observed difference is statistically contained within this margin. We also report paired \(t\)-tests to show whether any conventional difference is detected. For CIFAR-10, we evaluate all ten classes and replicate the full class sweep across three random seeds ($42$ primary, $1337$, $999$), yielding thirty class-seed runs per condition. We apply the same ten-class protocol to CIFAR-100. For each class, results are averaged over seeds before paired statistical testing, so that each class contributes one paired observation ($n{=}10$). The multi-seed runs are used to assess stability and to distinguish systematic mask effects from seed-dependent variation. Where the few-shot prototype metric is high-variance (CIFAR-100, per-class SD $\approx13$pp), a standard TOST power calculation ($\alpha{=}0.05$, power $0.8$) indicates roughly $n\approx40$ classes would be needed to resolve the comparison, four times our census; the lower-variance linear probe is the decisive metric there.

\section{Results}
\label{sec:results}

We now evaluate whether saliency-based weight selection changes what the model encodes after class unlearning. The results are organized around the three questions introduced in Section~\ref{sec:method}. We first test whether the saliency mask produces different representation-level outcomes from random or unconstrained updates. We then analyze why the three conditions behave similarly by inspecting the geometry of the learned representations and the distribution of the forgetting gradients. Finally, we examine whether the saliency mask itself contains class-specific information.

\subsection{Saliency-based masking does not yield distinguishable representation-level forgetting}
\label{sec:cifar10}

We begin with CIFAR-10, where class boundaries are sufficiently separated for prototype recovery to provide a discriminative representation-level test. Table~\ref{tab:rec} reports the mean prototype recovery across all ten forget classes and three random seeds. The three matched conditions (SalUn, random mask, and RL-full) produce nearly identical recovery values: $93.2\%$, $93.9\%$, and $93.6\%$, respectively. In contrast, the Gold retrain reaches $41.5\%$, while RandInit (the untrained texture floor) reaches $25.5\%$.

This pattern has two implications. First, all three approximate unlearning conditions leave the forgotten class substantially more recoverable than exact retraining. Second, and more importantly for our ablation, the saliency mask does not produce a representation-level outcome distinguishable from either a random mask of equal sparsity or an unconstrained update.

The paired equivalence tests in Table~\ref{tab:tost} confirm this interpretation. At the prototype recovery level, all three pairwise comparisons on CIFAR-10 fall within the pre-specified $\pm5$ percentage-point equivalence margin, with TOST $p<0.001$. The same conclusion holds for the linear-probe metric: all pairwise differences are at most $0.2$ percentage points and are statistically equivalent at the same margin. Thus, under the controlled conditions of our experiment, the specific identity of the saliency-selected weights does not explain representation-level class recoverability.

Importantly, this result does not arise because the unlearning objective failed at the output level. In all three conditions, forget accuracy drops to $0.0\%$, meaning that the model no longer predicts the forgotten class as its original label. The residual information is therefore not visible from output accuracy alone; it remains accessible through internal representations.

The result is also not specific to SalUn's selection rule. The Fisher-mask condition, which ranks weights by a second-moment importance statistic instead of gradient magnitude, lands at $93.4\%$ recovery on CIFAR-10 and $25.6\%$ on CIFAR-100, statistically equivalent to SalUn on both datasets (TOST $p{<}0.001$ and $p{=}0.01$ respectively; Table~\ref{tab:tost}) and to the random mask and RL-full on CIFAR-10 ($p{<}0.001$), with the same inconclusive prototype pattern on CIFAR-100 that \S\ref{sec:cifar100} documents for the other pairs. On the linear probe it is equivalent to all three conditions on both datasets ($p{<}0.001$). \S\ref{sec:specificity} shows why: the two criteria select nearly the same weights.

\begin{table}[!ht]
\small
\centering
\renewcommand{\arraystretch}{1.2}
\caption{\textbf{Representation-level recovery after unlearning.}
Mean prototype recovery ($k{=}5$) on CIFAR-10 and CIFAR-100: mean over 10 classes, each class averaged over 3 seeds; $\pm$ is the SD of that 10-class mean across the three seeds (per-class spreads are given in Table~\ref{tab:appendix_recovery}).
Higher values indicate that the forgotten class remains more recoverable from internal features.
Reference rows are shown in grey.}
\label{tab:rec}
\resizebox{\columnwidth}{!}{%
\begin{tabular}{lcc}
\toprule
Condition & CIFAR-10 & CIFAR-100 \\
\midrule
SalUn (saliency mask) & $93.2 \pm 0.3$ & $25.5 \pm 5.4$ \\
Random mask (same sparsity)    & $93.9 \pm 0.5$ & $26.7 \pm 4.8$ \\
RL-full (no mask)              & $93.6 \pm 0.8$ & $27.4 \pm 4.2$ \\
Fisher mask (2nd criterion)    & $93.4 \pm 0.2$ & $25.6 \pm 2.4$ \\
\midrule
\textcolor{gray}{Gold (retrain)}            & \textcolor{gray}{$41.5 \pm 0.1$} & \textcolor{gray}{$25.5 \pm 0.2$} \\
\textcolor{gray}{RandInit (texture floor)}  & \textcolor{gray}{$25.5 \pm 1.2$} & \textcolor{gray}{$22.4 \pm 1.5$} \\
\bottomrule
\end{tabular}}
\end{table}

\begin{table*}[t]
\centering
\caption{
Paired equivalence and difference tests with equivalence margin $\delta=\pm5$ percentage points.
Each paired observation corresponds to one forget class, averaged over three seeds.
For TOST, $p<0.05$ indicates statistical equivalence within the pre-specified margin.
Prototype recovery resolves the comparison clearly on CIFAR-10 but is high-variance on CIFAR-100.
Linear-probe recall provides a lower-variance representation-level test and confirms equivalence across all mask conditions on both datasets.
}
\label{tab:tost}
\small\setlength{\tabcolsep}{3.4pt}%
\renewcommand{\arraystretch}{1.1}
\begin{tabular}{lcccc}
\toprule
& \multicolumn{2}{c}{CIFAR-10} & \multicolumn{2}{c}{CIFAR-100} \\
\cmidrule(lr){2-3}\cmidrule(lr){4-5}
Comparison & diff (pp) & TOST $p$ & diff (pp) & TOST $p$ \\
\midrule
\multicolumn{5}{l}{\textit{Prototype recovery ($k{=}5$)}} \\
SalUn vs.\ RL-full \emph{(mask vs.\ no mask)} & $-0.4$ & $\mathbf{<0.001}$ & $-1.9$ & $0.20$ \\
Random mask vs.\ RL-full                               & $+0.3$ & $\mathbf{<0.001}$ & $-0.7$ & $0.17$ \\
SalUn vs.\ Random mask                        & $-0.6$ & $\mathbf{<0.001}$ & $-1.2$ & $0.08$ \\
Fisher mask vs.\ SalUn \emph{(2nd criterion)} & $+0.2$ & $\mathbf{<0.001}$ & $+0.1$ & $\mathbf{0.01}$ \\
\midrule
\multicolumn{5}{l}{\textit{Linear-probe recall (avgpool)}} \\
SalUn vs.\ RL-full                            & $+0.1$ & $\mathbf{<0.001}$ & $-0.6$ & $\mathbf{<0.001}$ \\
Random mask vs.\ RL-full                               & $+0.1$ & $\mathbf{<0.001}$ & $-0.1$ & $\mathbf{<0.001}$ \\
SalUn vs.\ Random mask                        & $-0.0$ & $\mathbf{<0.001}$ & $-0.4$ & $\mathbf{<0.001}$ \\
Fisher mask vs.\ SalUn                        & $-0.0$ & $\mathbf{<0.001}$ & $+0.1$ & $\mathbf{<0.001}$ \\
\midrule
\multicolumn{5}{l}{\textit{vs.\ Gold (paired difference test)}} \\
SalUn vs.\ Gold & \multicolumn{2}{c}{$+51.7$pp, $p<0.001$} & \multicolumn{2}{c}{$+0.0$pp, n.s.} \\
RL-full vs.\ Gold        & \multicolumn{2}{c}{$+52.1$pp, $p<0.001$} & \multicolumn{2}{c}{$+1.9$pp, n.s.} \\
Random mask vs.\ Gold    & \multicolumn{2}{c}{$+52.4$pp, $p<0.001$} & \multicolumn{2}{c}{$+1.2$pp, n.s.} \\
\bottomrule
\end{tabular}
\end{table*}

\subsection{CIFAR-100: fine-grained structure limits prototype-based conclusions}
\label{sec:cifar100}

CIFAR-100 presents a different evaluation regime. Because classes are finer-grained and more visually overlapping, prototype recovery is less discriminative than on CIFAR-10. This is visible in Table~\ref{tab:rec}: the Gold retrain itself reaches $25.5\%$ prototype recovery, essentially tied with the three approximate unlearning conditions ($25.5$--$27.4\%$) and only modestly above the RandInit texture floor ($22.4\%$). In other words, even a model retrained without the forgotten class can still yield features from which that class is partially recoverable.

This observation should not be interpreted as evidence that approximate unlearning is as effective as exact retraining on CIFAR-100. Rather, it indicates that few-shot prototype recovery is strongly influenced by the geometry of fine-grained visual classes. When classes share low-level or semantic structure, a nearest-centroid attack can recover part of the forgotten class even from features that were never trained on that class.

For this reason, prototype recovery does not provide a high-powered equivalence test on CIFAR-100. As shown in Table~\ref{tab:tost}, none of the prototype-based mask comparisons is statistically equivalent or significantly different. The observed differences are small in magnitude, but the per-class variance is too large for the test to resolve them reliably.

The linear probe provides a more stable representation-level comparison. On CIFAR-100, all three mask conditions are statistically equivalent under linear-probe recall, with pairwise differences of at most $0.6$ percentage points and TOST $p<0.001$. Thus, while prototype recovery is limited by the fine-grained structure of CIFAR-100, the lower-variance linear-probe metric supports the same conclusion obtained on CIFAR-10: saliency-based masking does not yield a distinguishable representation-level advantage over random or unconstrained updates.

The multi-seed analysis further supports this interpretation. Prototype recovery varies substantially across seeds on CIFAR-100, with several classes showing large seed-dependent swings. These fluctuations do not follow a consistent direction in favor of either the saliency mask or the random mask. We therefore interpret CIFAR-100 as evidence that fine-grained datasets require care in representation-level auditing: prototype recovery can expose residual structure, but it may not always provide a sufficiently stable test for comparing closely related unlearning mechanisms.

\subsection{Random masks behave like saliency masks across classes and sparsity levels}
\label{sec:random}

The most direct test of the saliency hypothesis is the comparison between the SalUn mask and a random mask with the same sparsity. If the saliency criterion identifies parameters that are specifically responsible for forgetting a class, then replacing those parameters with a random subset of equal size should degrade representation-level forgetting.

This is not what we observe. On CIFAR-10, SalUn and random masking are statistically equivalent under prototype recovery, with a mean difference of $-0.6$ percentage points and TOST $p<0.001$. They are also equivalent under linear-probe recall, where the difference is effectively zero. On CIFAR-100, the prototype comparison remains statistically indeterminate because of the high per-class variance discussed above, but the linear probe again shows equivalence, with a difference of $-0.4$ percentage points and TOST $p<0.001$.

\begin{figure*}[t]
\centering
\includegraphics[width=0.85\textwidth]{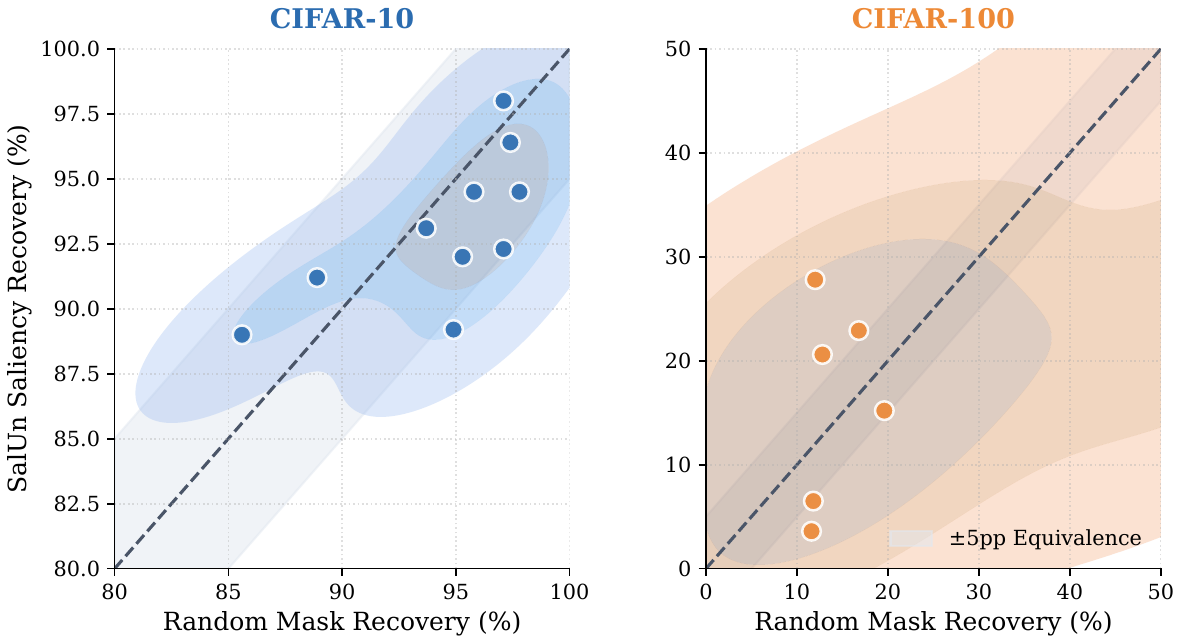}
\caption{\textbf{Saliency and random masks yield similar class-level recovery.}
Prototype recovery under the SalUn saliency mask and a random mask of equal sparsity.
Each point corresponds to one forget class; the shaded region denotes the $\pm5$ percentage-point equivalence band.
On CIFAR-10, nearly all classes fall within the equivalence region.
On CIFAR-100, per-class differences are larger but do not show a consistent direction across seeds.}
\label{fig:perclass}
\end{figure*}

Figure~\ref{fig:perclass} shows the class-level comparison. On CIFAR-10, nine of ten classes fall within the $\pm5$ percentage-point equivalence band. The only marginal outlier is \textit{frog}, where the saliency mask trails the random mask by $5.7$ percentage points at the primary seed. This deviation does not generalize across the remaining classes or seeds. On CIFAR-100, individual class differences are larger, but their signs change across seeds, indicating seed-dependent variability rather than a systematic advantage of one mask over the other.

We also test whether this conclusion depends on the sparsity level. A sweep over sparsity ratios in $\{0.1,0.3,0.5,0.7\}$ on three representative classes shows that, on CIFAR-10, the SalUn and random-mask outcomes remain close across sparsity levels, with small differences of mixed sign; on CIFAR-100 the variability is larger, but no sparsity level yields a consistent advantage for the saliency mask. Table~\ref{tab:maskgeom}(b) in \ref{app:tables} extends this check from three classes to the full ten-class census at every ratio, using the mask's raw overlap with an independent random mask rather than the noisier downstream recovery outcome: the overlap matches the analytic chance-level null $r/(2-r)$ to within $0.0005$ at every ratio on both datasets, confirming the same conclusion with a cleaner, complete measurement.

Together, these results suggest that the representation-level behavior of SalUn is not explained by the specific identity of the saliency-selected weights. The outcome is instead consistent with a setting in which the objective and the gradient structure already constrain where the update can act, leaving limited room for the precise mask identity to change the recovered representation.

\subsection{Mask identity does not change the redirection pattern}
\label{sec:redirection_results}

The previous results show that the amount of recoverable class information is similar across masks. A remaining possibility is that different masks preserve the same amount of information but redirect forgotten samples toward different retained classes. To test this, we record the most frequent top-1 prediction assigned to held-out forget-class samples after unlearning.

Table~\ref{tab:redirection} in \ref{app:tables} summarizes this analysis. On CIFAR-10, all three unlearning conditions redirect each forgotten class toward the same dominant wrong class. On CIFAR-100, the agreement is slightly weaker, as expected from the finer-grained class structure, but the three conditions still agree for most classes and differ mainly when multiple candidate classes receive similar fractions of predictions.

This result complements the recovery analysis. The saliency mask not only preserves a similar amount of representation-level information as the random and full-update baselines, but it also tends to move forgotten samples toward the same neighboring classes. The redirection pattern appears to be governed more by the geometry of the learned feature space than by mask-specific parameter selection.

\subsection{Gradient concentration explains the equivalence between masks}
\label{sec:mechanism}

The matched ablation shows that saliency, random, and unconstrained updates produce similar representation-level outcomes. We now ask why. The central observation is that the unlearning gradients are already highly localized before any mask is applied.

\begin{figure}[t]
\centering
\includegraphics[width=0.98\linewidth]{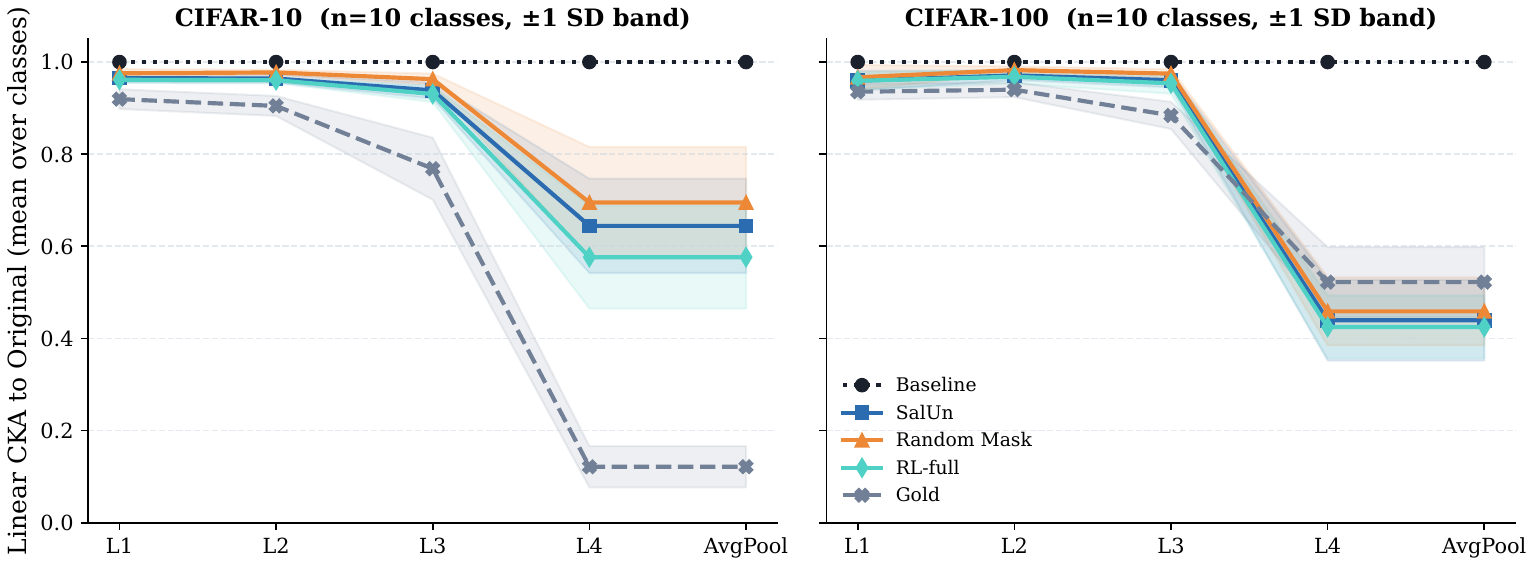}
\caption{\textbf{All matched conditions reshape the same late representation.}
Layer-wise linear CKA between each unlearned model and the original model, averaged over ten classes.
Early and intermediate layers remain close to the original representation, while the final residual block changes substantially.
The SalUn, random-mask, and RL-full profiles are closely aligned, indicating that the three updates affect similar representational regions.}
\label{fig:cka}
\end{figure}

Figure~\ref{fig:cka} reports the layer-wise CKA between the original model and the unlearned models. Across both datasets, the early and intermediate layers remain largely unchanged after unlearning. Layers 1 through 3 maintain high similarity to the original model, while most of the representational change is concentrated in the final residual block. This pattern is nearly identical for the SalUn mask, the random mask, and the unconstrained update.

\begin{figure}[t]
\centering
\includegraphics[width=0.98\linewidth]{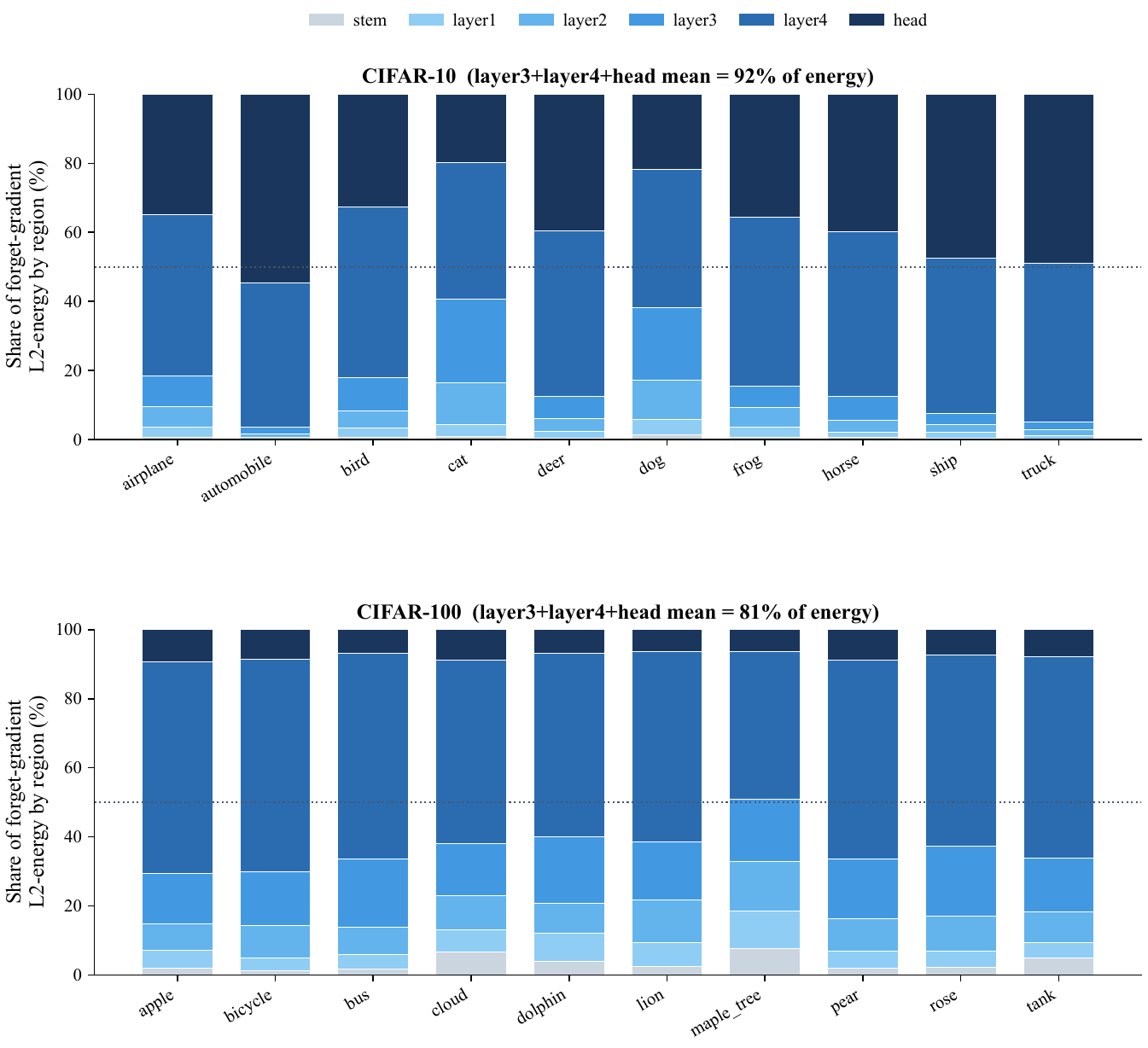}
\caption{\textbf{Forget gradients are concentrated in late network regions before masking.}
Share of the forget-loss gradient's squared $\ell_2$-norm by network region, measured before any unlearning step.
Layer 3, layer 4, and the classifier head account for most of the gradient energy on both datasets.
As a result, different masks are applied to gradients that already concentrate their update direction in the same late subspace.}
\label{fig:gradenergy}
\end{figure}

The gradient-energy analysis in Figure~\ref{fig:gradenergy} explains this behavior. Before any mask is applied, the forget-loss gradient is already concentrated in the later parts of the network. On CIFAR-10, layer 3, layer 4, and the classifier head together account for approximately $92\%$ of the squared gradient norm (three-seed mean; per-seed range ${\approx}87$--$99\%$). On CIFAR-100, they account for approximately $81\%$ (range ${\approx}77$--$88\%$). Within this late region the energy is dominated by layer 4 and the head, while layer 3 contributes only ${\approx}9\%$ (CIFAR-10); this is why layer 3 can carry gradient energy yet retain CKA${\ge}0.9$ to the original. The random-label gradient that SalUn actually optimizes is even more concentrated in the late layers (${\approx}99\%$ on CIFAR-10, ${\approx}89\%$ on CIFAR-100). We measure this concentration directly rather than assume it: the retain-set gradient shows the same late concentration (${\approx}95\%$ / ${\approx}79\%$), confirming it is a generic property of the converged network's loss geometry rather than something the forget set induces. It is consistent with the broader finding that parameter importance is highly non-uniform \citep{frankle2019lottery,han2015learning,hoefler2021sparsity} and with the geometry of neural collapse \citep{papyan2020neural} invoked by Gao et al. \citep{gao2026illusion}, though the \emph{late-layer} localization is a property of class-forget gradients that we establish empirically here.

This provides a mechanistic explanation for the equivalence observed above. The mask determines which coordinates of the gradient are allowed to update the model, but the gradient itself already determines where most update energy lies. Therefore, different masks are applied to update directions that are structurally concentrated in the same late network regions. This limits the ability of mask identity alone to produce substantially different representation-level outcomes.

This finding should be interpreted as an empirical mechanism rather than a general theorem. We do not claim that gradient concentration must occur in all architectures, datasets, or unlearning settings. Rather, in the ResNet-18 class-unlearning setting studied here, the observed gradient distribution provides a direct explanation for why SalUn, random masking, and unconstrained random-label updates reshape similar parts of the representation and leave similar amounts of class information recoverable.

\subsection{The forgotten class remains geometrically coherent}
\label{sec:cluster_survival}

CKA identifies where the representation changes, but it does not directly indicate whether the forgotten class remains organized as a coherent group in feature space. To assess this, we measure the cosine silhouette of the forget class in the penultimate representation.

Table~\ref{tab:silhouette} in \ref{app:tables} shows that the SalUn, random-mask, RL-full, and Fisher conditions all preserve a high forget-class silhouette on CIFAR-10, in the same range as the original trained model (${\approx}0.6$--$0.7$). In contrast, the Gold retrain pushes the silhouette toward zero, indicating that the forgotten class no longer forms the same coherent cluster in the feature space of a model trained without that class.

\begin{figure*}[t]
\centering
\includegraphics[width=\textwidth]{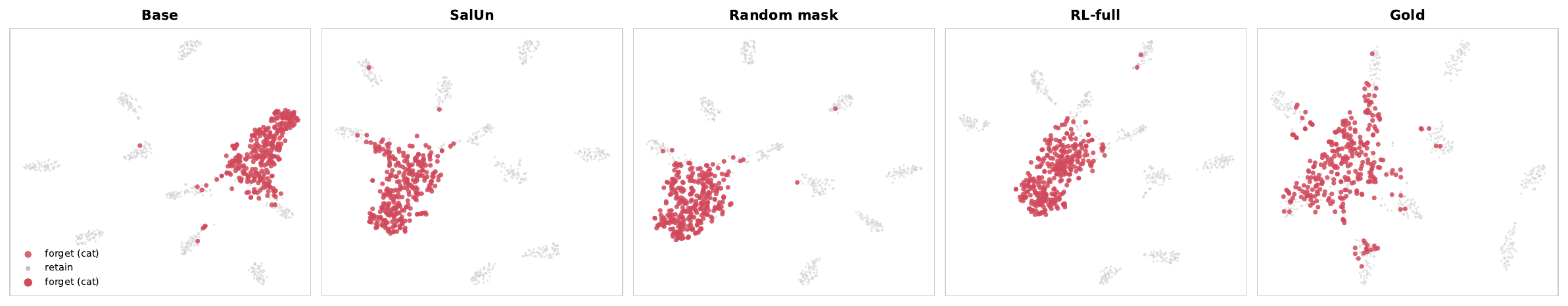}
\caption{\textbf{Feature-space visualization of class survival.}
t-SNE of penultimate features for CIFAR-10 class 3 (\textit{cat}) against all retained classes.
Under SalUn, random mask, and RL-full, the forget-class samples remain compact and separated from most retained samples.
Under Gold, the cluster is less coherent and partially dispersed among neighboring retained classes.}
\label{fig:tsne}
\end{figure*}

The same qualitative pattern is visible in the t-SNE visualization in Figure~\ref{fig:tsne}. For the representative CIFAR-10 class \textit{cat}, the forget-class samples remain compact and visually separated after SalUn, random masking, and unconstrained random-label updates. Under Gold, the cluster becomes less compact and partially overlaps with neighboring retained classes.

These results clarify the difference between output-level forgetting and representation-level erasure. Although all unlearning conditions suppress the forgotten class at the classifier output, they do not dissolve the internal feature-space structure that makes the class recoverable.

\subsection{Saliency masks show limited class specificity}
\label{sec:specificity}

The previous analysis shows that different masks act on gradients concentrated in similar late-layer regions. We now examine whether the saliency masks themselves encode class-specific information. If SalUn identifies parameters that are specific to a particular forgotten class, masks computed for different forget classes should exhibit substantially lower overlap than masks computed from the same or related data distributions.

For each dataset, class, and seed, we compute the SalUn forget-class mask and compare it with three alternatives: masks computed from other forget classes, masks computed from the retain set, and masks computed from random labels on the same forget images. We measure overlap using the Jaccard index.

\begin{figure}[t]
\centering
\includegraphics[width=\linewidth]{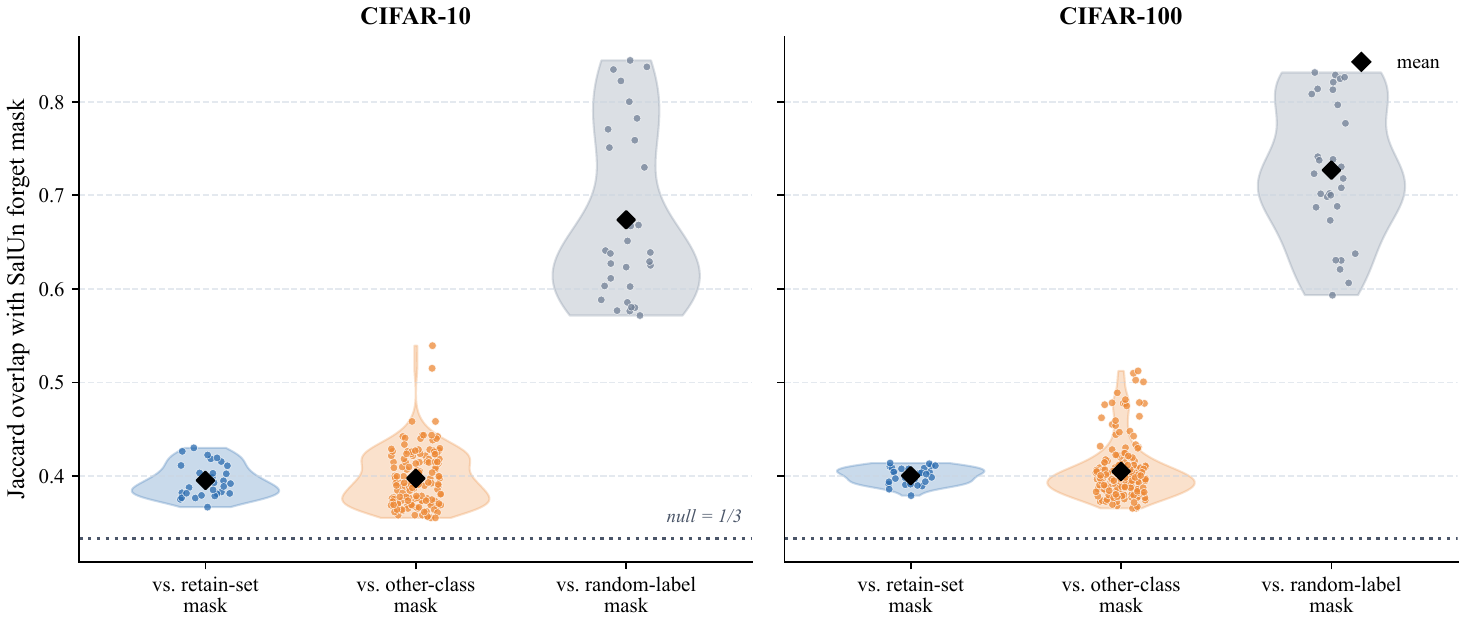}
\caption{\textbf{Saliency masks show limited class specificity.}
Jaccard overlap between the forget-class SalUn mask and three comparator masks over class-seed combinations.
Other-class and retain-set masks exhibit very similar overlap with the forget-class mask, suggesting that the selected parameters are not strongly specific to the forgotten class.}
\label{fig:maskspec}
\end{figure}

Table~\ref{tab:specificity} and Figure~\ref{fig:maskspec} show that masks computed from other classes overlap with the forget-class mask almost as much as masks computed from the retain set. On CIFAR-10, the mean Jaccard overlap is $0.397$ for other-class masks and $0.395$ for retain-set masks. On CIFAR-100, the corresponding values are $0.405$ and $0.400$. This indicates limited class specificity: the saliency mask is not substantially more aligned with the forgotten class than with gradients computed from retained data.

We summarize this as a \emph{specificity index}, $(J_{\mathrm{other}}-J_{0})/(1-J_{0})$: the mask's excess overlap above the permutation null $J_{0}{=}1/3$ (for $50\%$-sparsity masks), expressed as a fraction of the maximum possible excess. It is $0.096$ on CIFAR-10 and $0.107$ on CIFAR-100, with per-seed values spanning $0.09$--$0.11$ (the index quoted in the abstract and conclusion). Because the other-class overlap is indistinguishable from the retain-set overlap, the mask carries essentially zero forget-specific signal beyond what the architecture forces. This is the upstream reason a random mask of equal sparsity matches SalUn (\S\ref{sec:cifar10},~\S\ref{sec:random}): the weights SalUn selects to forget a class are largely those it would select for any class, or from no particular class at all. The practical implication is that SalUn's weight selection is effectively class-agnostic at the representation level, so privacy guarantees attributed to the saliency mechanism should not be assumed to follow from it.

The random-label comparator has higher overlap, with mean Jaccard values of $0.674$ on CIFAR-10 and $0.727$ on CIFAR-100. This comparison must be interpreted carefully because it reuses the same input images and therefore mixes label effects with shared activation patterns. We therefore do not treat it as a clean label-only ablation. Nevertheless, the result is consistent with the broader picture that saliency masks are strongly influenced by input-induced gradient structure.

The Fisher-mask condition closes the loop between mask geometry and outcome. Although it ranks weights by a different statistic (second moment rather than magnitude of the mean gradient), its mask overlaps SalUn's at $J{=}0.83$ (CIFAR-10) and $J{=}0.94$ (CIFAR-100) against the $1/3$ chance null, with ${\approx}89\%$ of its budget in layer\,3$+$4$+$head on both datasets (Table~\ref{tab:maskgeom}(c)). Gradient-derived selection rules do not merely produce the same unlearning outcome (Table~\ref{tab:rec}); they select nearly the same weights, because both statistics inherit the same late-concentrated gradient (\S\ref{sec:mechanism}). The class-agnosticism finding therefore covers the rule family, not one instance of it.

\begin{table}[!ht]
\centering
\caption{Mask overlap across ten forget classes and three seeds per dataset. The retain-set row provides an empirical baseline for non-forget-specific gradient structure. The random-label row reuses the same forget images with scrambled labels and should therefore be interpreted as reflecting both label and input-activation effects.}
\label{tab:specificity}
\renewcommand{\arraystretch}{1.15}
\begin{tabular}{lcc}
\toprule
Comparator & CIFAR-10 mean $J$ & CIFAR-100 mean $J$\\
\midrule
Other class  & $0.397$ & $0.405$ \\
Retain set   & $0.395$ & $0.400$ \\
Random label & $0.674$ & $0.727$ \\
\bottomrule
\end{tabular}
\end{table}

As a preliminary diagnostic, we also tested a contrastive mask based on the difference between average forget and retain gradients for one class and one seed. This construction did not materially increase class specificity and further concentrated the selected parameters in the final residual block. Because this analysis is not replicated across all classes and seeds, we treat it only as a supporting diagnostic rather than as a main result.

Overall, the mask-overlap analysis supports the interpretation developed above. Saliency-based masking does not appear to isolate a strongly class-specific subset of parameters in this setting. Instead, the selected coordinates largely reflect gradient structures that are shared across classes and concentrated in the same late network regions.

\subsection{Causal test: forgetting without the late subspace}
\label{sec:causal}

The mechanism account of \S\ref{sec:mechanism} is correlational: every gradient-derived
mask lands in the late subspace, and every one produces the same outcome. To test the
account causally, the early-mask condition (\S\ref{sec:method}) inverts the placement:
all of stem$+$layers\,1--3 is trainable ($24.8\%$ of parameters) and layer\,4 plus the
head are held bit-exact frozen, so the update is confined to the subspace that no
gradient-derived rule selects. A forced variant repeats this at $3\times$ the learning
rate under gradient clipping. Both run the identical objective, schedule, and protocol.

\paragraph{Output-level forgetting does not need the late subspace} Across all $60$
class-seed runs on the two datasets, the early mask drives forget accuracy to
$0$--$4\%$, with retain accuracy at the Gold level ($72$--$74\%$ on CIFAR-100).
Updating early weights changes late-layer \emph{activations} without touching late
\emph{weights}, and the random-label objective exploits this route without difficulty.
The forget loss can therefore be satisfied from essentially anywhere in the network;
reaching the late subspace is not the binding constraint on output-level forgetting.
Notably, the early mask needs more optimization to get there (median $2$--$3$ repair
epochs to convergence on CIFAR-100, against $1$ for the saliency-family masks), which
rules out under-training as an explanation for the representational differences below.

\paragraph{What placement does control: prototype geometry, not separability}
The two representation metrics dissociate cleanly. On the \emph{linear probe}, the early
mask is statistically equivalent to SalUn on both datasets (recall $96.8$ vs.\
$98.0$\% on CIFAR-10, $92.2$ vs.\ $92.5$\% on CIFAR-100; TOST $p{<}0.001$ for both) and
sits ${\approx}5$pp above Gold ($t$-test $p{<}10^{-4}$ on both datasets): linear
separability of the forget class survives \emph{regardless of where the update lands},
extending the equivalence result of \S\ref{sec:cifar10}--\S\ref{sec:random} causally.
On \emph{prototype recovery}, placement matters where the metric has room to move: on
CIFAR-100 the early mask leaves recovery at $59.0\pm1.4$\%, close to the base model's
$67.9$\% and $+33.5$pp above SalUn (paired $t$-test $p{<}10^{-5}$; TOST
$p{\approx}1$), while every late-subspace condition sits at the Gold floor
($25$--$27$\%). On CIFAR-10 the prototype metric is ceiling-saturated (base $94.3$\%)
and the early mask is TOST-equivalent to SalUn ($92.1$ vs.\ $93.2$\%,
$p{=}0.003$). The forget-class silhouette completes the picture: under the early mask it
drops to $0.56$ (forced: $0.53$) against $0.61$--$0.68$ for the late masks, yet remains
far above Gold's $-0.09$; early updates \emph{distort} the cluster's geometry more while
\emph{dissolving} it less.

\begin{table}[!ht]
\centering
\renewcommand{\arraystretch}{1.15}
\caption{\textbf{The causal placement test.} Forget accuracy, linear-probe recall,
prototype recovery ($k{=}5$), and CIFAR-10 forget-class silhouette for the early-only
mask (layer\,4$+$head frozen) and its forced ($3\times$ LR) variant, against
SalUn and Gold. All values seed-averaged over the $10$-class census. Late
updates move the geometry-sensitive prototype metric to the Gold floor on CIFAR-100
without reducing probe recall; early updates achieve the same output-level forgetting
while moving neither.}
\label{tab:causal}
\resizebox{\columnwidth}{!}{%
\begin{tabular}{lcccccc}
\toprule
& & \multicolumn{2}{c}{Probe recall (\%)} & \multicolumn{2}{c}{Recovery (\%)} & Silh.\ \\
\cmidrule(lr){3-4}\cmidrule(lr){5-6}
Condition & fa (\%) & C-10 & C-100 & C-10 & C-100 & C-10 \\
\midrule
Early mask            & $0$--$4$ & $96.8$ & $92.2$ & $92.1$ & $59.0$ & $0.56$ \\
Early mask, $3\times$ & $0$--$4$ & $96.4$ & $91.2$ & $90.4$ & $59.4$ & $0.53$ \\
\midrule
\textcolor{gray}{SalUn} & \textcolor{gray}{$0$} & \textcolor{gray}{$98.0$} & \textcolor{gray}{$92.5$} & \textcolor{gray}{$93.2$} & \textcolor{gray}{$25.5$} & \textcolor{gray}{$0.62$} \\
\textcolor{gray}{Gold}           & \textcolor{gray}{$0$} & \textcolor{gray}{$91.8$} & \textcolor{gray}{$87.4$} & \textcolor{gray}{$41.5$} & \textcolor{gray}{$25.4$} & \textcolor{gray}{$-0.09$} \\
\bottomrule
\end{tabular}}
\end{table}

\paragraph{Reading} The intervention refines the mechanism account in one direction and
falsifies it in another. Falsified: the late subspace is \emph{not} necessary for
output-level forgetting, so gradient concentration does not gate whether a mask can
forget. Refined: what late-subspace updates specifically produce is the displacement of
\emph{prototype geometry} toward Gold-like values, and this displacement is achievable
\emph{only} from the late subspace, without ever constituting erasure, since the linear
probe stays at base level under every condition. Geometry-sensitive scores can therefore
be driven toward the exact-retrain reference by any late-subspace update while the class
remains fully separable, and a placement that avoids the late subspace forgoes even that
appearance. Two caveats: the forced variant pays $3$--$6$pp of retain accuracy on
CIFAR-100 (below our utility gate on all $30$ runs), so we treat it as supporting
evidence only; and $2$ of $30$ forced runs did not reach the forget-accuracy criterion.

\FloatBarrier

\section{Discussion}
\label{sec:discussion}

The results provide a consistent answer to the question posed in this paper: in the class-unlearning setting studied here, the identity of the saliency-selected weights does not explain representation-level forgetting. Saliency masking, random masking, and unconstrained random-label updates lead to statistically equivalent representation-level outcomes under the most stable metrics. The mechanism behind this equivalence is not that masks have no effect in general, but that the gradients used for unlearning are already strongly concentrated in late network regions before masking is applied. As a result, different masks operate on update directions that are structurally biased toward the same representational subspace.

This interpretation reframes the role of saliency-based masking. SalUn-style masks may still provide computational or optimization advantages in some settings, and our results do not challenge their reported output-level or generative behavior. However, for representation-level class forgetting on CIFAR-10/100 with ResNet-18, the saliency criterion does not appear to isolate a class-specific subset of parameters that determines whether the forgotten class remains recoverable. Instead, recoverability is better explained by the geometry of the learned representation and by the concentration of the unlearning gradient.

\subsection{Implications for unlearning evaluation}

A central implication is that output-level forgetting is not sufficient to assess deletion quality. In our experiments, all three unlearning conditions reduce forget accuracy to zero, yet the forgotten class remains highly recoverable from internal features on CIFAR-10. This gap illustrates why representation-level auditing is necessary when unlearning is used to support privacy, deletion, or compliance claims.

The results also suggest that different representation-level metrics should be interpreted with care. Prototype recovery is useful because it directly tests whether a class remains clustered in feature space. However, on fine-grained datasets such as CIFAR-100, prototype recovery can be strongly affected by shared visual structure among classes. In this setting, even the Gold retrain can yield non-trivial prototype recovery, meaning that the metric reflects both residual model information and the intrinsic geometry of the dataset. Linear probing provides a lower-variance complementary test and, in our experiments, gives the clearest comparison between masking conditions.

Taken together, these findings support a broader evaluation principle: unlearning should be audited at multiple levels. Output accuracy verifies whether the classifier still predicts the forgotten label; representation-level metrics test whether the information remains accessible inside the model; and reference models such as Gold and RandInit help separate retained class information from dataset-level visual structure.

\subsection{Implications for method design}

The main design implication is that better weight selection alone may not be sufficient for representation-level forgetting. If the objective produces gradients concentrated in late layers, then any binary mask applied to those gradients will inherit the same structural bias. Under this condition, changing the identity of selected weights can have limited effect on the geometry of earlier or intermediate representations, where class information may still remain organized.

This interpretation is aligned with recent representation-level unlearning methods that operate directly in latent space. Projection-based \citep{pour2025representation}, contrastive \citep{tang2026clreg}, and intermediate-layer \citep{lee2026erase} objectives explicitly target the structure of learned representations rather than relying only on parameter selection. Our results provide a mechanistic motivation for this direction: if the goal is to dissolve the internal structure that makes a forgotten class recoverable, then the objective should act on that structure directly.

This does not imply that masking is useless. A mask can still constrain optimization, reduce computational cost, or protect parts of the model from unnecessary updates. The narrower conclusion is that, in our experiments, the SalUn saliency mask is not the factor that determines representation-level class erasure. Future unlearning methods may need to combine parameter-efficient updates with explicit representation-level objectives if they aim to provide stronger deletion semantics.

The causal placement test (\S\ref{sec:causal}) sharpens this principle into a stronger form. A mask that excludes the late subspace entirely still drives forget accuracy to zero, so gradient concentration does not gate whether a masked update can forget; what it gates is the displacement of prototype geometry toward Gold-like values, which only late-subspace updates produce and which never amounts to erasure, since linear separability survives at base level under every placement. No placement of a logit-level update erases the class; late placements merely make geometry-sensitive metrics \emph{look} more like exact retraining. A weight-space route to representation-level forgetting would therefore have to change what the objective optimizes, not where the update lands.

\subsection{Limitations}

Our conclusions are bounded by the experimental setting. We study class unlearning on CIFAR-10 and CIFAR-100 using ResNet-18. This setting is important because it matches the closest representation-level auditing literature and allows controlled class-wise evaluation, but it does not cover all unlearning regimes. In particular, we do not study random data-point forgetting, large-scale datasets, transformer architectures, multimodal models, large language models, or diffusion models \citep{gandikota2023erasing}.

The interpretation of CIFAR-100 also requires caution. Because CIFAR-100 contains fine-grained and visually related classes, prototype recovery is less discriminative than on CIFAR-10. The fact that Gold and RandInit can exhibit non-trivial recovery does not mean that exact retraining fails in a general sense. Rather, it indicates that nearest-centroid recovery can exploit shared visual structure even when the target class was not present during training. For this reason, our strongest CIFAR-100 conclusions rely on the lower-variance linear-probe comparisons.

Another limitation concerns the mask-specificity analysis. We use Jaccard overlap to quantify whether masks computed from different classes select similar parameter subsets. This is an informative structural diagnostic, but it is not a complete measure of functional equivalence. Two masks can have similar overlap yet still interact differently with the gradient magnitudes or with the local loss landscape. Our functional ablation addresses this point empirically, but a more complete theory of mask function remains open.

Finally, the gradient-concentration mechanism is empirical. We show that it explains the behavior observed in the studied setting, but we do not prove that such concentration must arise in all architectures or datasets. Architectures with different inductive biases, such as vision transformers or large foundation models, may distribute forget-gradient energy differently. Testing whether the same mechanism holds at scale is an important direction for future work.

\subsection{Future directions}

A first direction is to replicate the matched-compute ablation beyond CIFAR-scale convolutional networks. ImageNet-scale classifiers, vision transformers, and multimodal encoders would clarify whether late-layer gradient concentration is a general phenomenon or a property of the ResNet-18 class-unlearning setting.

A second direction is to develop stronger representation-level objectives. The present results suggest that objectives acting only through output labels may fail to dissolve class structure in feature space. Future methods should therefore explore losses that explicitly reduce forget-class separability, reshape class prototypes, or align forgotten samples with semantically related retained classes while preserving utility.

A third direction is methodological. Representation-level unlearning evaluation would benefit from standardized protocols combining linear probes, prototype recovery, CKA-style similarity analysis, membership inference, and carefully designed reference models. Such protocols would make it easier to distinguish genuine deletion from superficial output forgetting and to compare results across datasets with different geometric properties.

Overall, our findings suggest that the next step for machine unlearning is not simply to design more selective masks, but to understand and control the representation geometry that makes forgotten information recoverable.

\section{Conclusion}
\label{sec:conclusion}

This paper investigated the role of saliency-based weight selection in representation-level class unlearning. Using a matched-compute ablation, we isolated the masking mechanism used by SalUn from the random-label objective on which it operates. Across CIFAR-10 and CIFAR-100 with ResNet-18, we found that saliency masks, random masks of equal sparsity, a diagonal-Fisher mask, and unconstrained random-label updates produce statistically equivalent representation-level behavior under the most stable evaluation metrics.

The main conclusion is not that masking is irrelevant in all unlearning settings. Rather, in the setting studied here, the specific identity of the saliency-selected weights does not appear to determine whether the forgotten class remains recoverable from internal representations. The observed equivalence is better explained by two empirical mechanisms: the unlearning gradients are already strongly concentrated in late network regions before masking is applied (${\approx}92\%$ of the squared gradient energy on CIFAR-10), and the resulting saliency masks show limited class specificity across forget classes (specificity index $0.09$--$0.11$; the two gradient-derived criteria we test overlap at Jaccard $0.83$--$0.94$). A causal placement intervention completes the picture: a mask forcibly excluded from the late subspace still drives forget accuracy to zero on every run, while linear separability of the class survives at base level under \emph{every} placement, so what late-subspace updates uniquely produce is Gold-like prototype geometry, not erasure.

These findings help clarify why output-level forgetting can be misleading. All matched unlearning conditions successfully suppress the forgotten class at the classifier output, yet the class remains recoverable from internal features. This indicates that reducing forget accuracy is not sufficient to establish representation-level erasure. Reliable unlearning evaluation should therefore combine output metrics with representation-level probes, feature-space analyses, and appropriate reference models such as exact retraining and random initialization.

More broadly, the results support a growing shift in machine unlearning research from parameter selection toward representation-level objectives, and concurrent work \citep{pour2025representation,tang2026clreg,lee2026erase} points the same way, consistent with the geometry of neural collapse \citep{papyan2020neural}. If forgotten information remains encoded as a coherent structure in feature space, then stronger deletion semantics may require objectives that act directly on that structure rather than relying only on increasingly selective masks. Future work should test whether the gradient-concentration mechanism observed here extends to larger datasets, transformer architectures, multimodal models, and other unlearning regimes.

\subsubsection*{Broader Impact Statement}

Machine unlearning is increasingly discussed as a technical mechanism for supporting data deletion, privacy, and regulatory compliance. Our results highlight a potential gap between apparent output-level forgetting and deeper representation-level removal. A method may stop predicting the forgotten class while still retaining internal features from which that class can be recovered. This does not introduce a new attack capability against deployed systems; rather, it provides a diagnostic perspective for evaluating claims about unlearning effectiveness.

The broader implication is that practitioners should be cautious when interpreting low forget accuracy as evidence of deletion. For applications where unlearning is used to support privacy-sensitive or compliance-critical decisions, representation-level auditing should complement standard output-level metrics. The causal test sharpens this warning: geometry-sensitive scores can be driven toward retrain-like values by any late-subspace update without any loss of class separability (\S\ref{sec:causal}), so audits that rely on them can be satisfied by updates that erase nothing. At the same time, the scope of our findings is limited to controlled class-unlearning experiments on CIFAR-scale vision models, and they should not be generalized to all architectures or domains without further validation.

%%, - CRediT author statement,-

\section*{CRediT authorship contribution statement}

\textbf{Billel Habbati:} Conceptualization, Methodology, Software, Formal analysis,
Investigation, Data curation, Writing -- original draft, Visualization.

\textbf{Alessio Merlo:}  Project administration, Validation, Writing -- review \& editing.

\textbf{Luca Verderame:} Supervision, Validation, Writing -- review \& editing.

\textbf{Meriem Guerar:} Supervision, Validation, Writing -- review \& editing.

%%, - AI use declaration,-

\section*{Declaration of generative AI and AI-assisted technologies in the manuscript preparation process}
During the preparation of this work the authors used Claude by Anthropic in order to improve the language and readability of selected passages. After using this tool, the authors reviewed and edited the content as needed and take full responsibility for the content of the published article.

%%, - Declaration of competing interest,-

\section*{Declaration of competing interest}

The authors declare that they have no known competing financial interests or personal relationships that could have appeared to influence the work reported in this paper.

%%, - Data availability,-

\section*{Data availability}

The code and experimental results needed to reproduce our findings are archived on Zenodo \citep{habbati2026code} at \url{https://doi.org/10.5281/zenodo.21507241} and are maintained at \url{https://github.com/neparino/Salun-Unlearning-Studies}. CIFAR-10 and CIFAR-100 are publicly available benchmark datasets.
%%, - Funding,-

\section*{Funding}

This research did not receive any specific grant from funding agencies in the public, commercial, or not-for-profit sectors. Computational resources were provided by the Computer Security Laboratory servers at DIBRIS, University of Genova.

\clearpage
%%, - bibliography,-
\bibliographystyle{elsarticle-harv}
\bibliography{references}

\clearpage
%%, - appendix,-

\appendix
\setcounter{table}{0}
\setcounter{figure}{0}

%% ============================================================================
%% Appendix, shared by the one-column and two-column builds. Starred floats
%% (table*/figure*) span both columns in two-column mode and behave as ordinary
%% floats in one-column mode. All floats are grouped in the final section so that
%% none strand text or leave a page mostly empty.
%% ============================================================================

This appendix has one purpose: to let every quantitative claim in
\S\ref{sec:cifar10}--\S\ref{sec:specificity} be checked against the exact
configuration and data that produced it, rather than taken on trust. It
follows the paper's own logic rather than a generic template.
\ref{app:repro} gives the configuration needed to reproduce any single run.
\ref{app:data} gives the complete data behind every reported average: full
per-class, per-seed recovery, CKA, and specificity numbers underlying RQ1
(\S\ref{sec:cifar10}--\S\ref{sec:random}), and the mask's own depth profile
and overlap with an independent random mask underlying RQ2--RQ3
(\S\ref{sec:mechanism}--\S\ref{sec:specificity}). \ref{app:tables} collects
every supplementary figure and table named above; all floats are grouped at
the end so none of them interrupts the running text or strands a stub of
text in an otherwise empty column.

\section{Implementation and Reproducibility Details}
\label{app:repro}

All three unlearning conditions share one base model and one evaluation
pipeline; only the mask $\mathbf{m}$ in Eq.~\ref{eq:mask} varies. The five
subsections below give the choices that carry the claim: how the base and
reference models are trained (\S A.1), why the three conditions differ only
in the mask (\S A.2), how representation survival is attacked and measured
(\S A.3), how equivalence is tested and what statistical power that test has
(\S A.4), and the compute and software needed to regenerate every number
(\S A.5).

\paragraph{A.1\quad Base and reference models} Each forget class gets its own base
model, an original network trained on the full dataset $\mathcal{D}$ for $182$
epochs, and its own \textbf{Gold} retrain on $\mathcal{D}_r$ (the same class removed)
for $40$ epochs. Both use AdamW with a cosine schedule; the base warms up over five
epochs, Gold over three. \textbf{RandInit} is an untrained network with the same
architecture, providing the per-class texture floor. Because Gold and RandInit are
computed per class, every recovery number in Table~\ref{tab:rec} and
Table~\ref{tab:appendix_recovery} is referenced against a matched exact-unlearning
baseline rather than a single global one.

\paragraph{A.2\quad The three conditions differ only in $\mathbf{m}$} All three
unlearning conditions start from the same base checkpoint and run the identical
random-label objective (Eq.~\ref{eq:rl}) with SGD (momentum $0.9$, learning rate
$0.013$, weight decay $5\times10^{-4}$) for ten repair epochs. SalUn applies
the saliency mask $\mathbf{m}_S$ (top-$50\%$ weights by forget-gradient magnitude);
\textbf{Random mask} applies $\mathbf{m}_R$ of equal $\ell_0$; \textbf{RL-full} sets
$\mathbf{m}=\mathbf{1}$. To keep the mask faithful, masked-out coordinates are held at
their base value \emph{and} their momentum buffer is zeroed each step, preventing
momentum from re-introducing updates the mask is meant to block. We adopt
SalUn's released CIFAR-10 hyperparameters unchanged on both datasets, so no
value is tuned to our result. The two diagnostic conditions reuse this exact loop and
differ only in mask construction: the \textbf{Fisher mask} ranks weights by diagonal
empirical Fisher importance (mean over batches of the squared forget-CE gradient) and
applies the same global top-$50\%$ threshold; the \textbf{Early mask} sets
$\mathbf{m}{=}1$ on the stem and residual blocks 1--3 and $\mathbf{m}{=}0$ on block 4
and the head ($24.8\%$ of parameters trainable), with the frozen coordinates pinned
identically. Its forced variant multiplies the learning rate by $3$ under a global
gradient-norm clip of $5$.

\paragraph{A.3\quad Attacks and metrics} Representation survival is measured with the
five tools of \S\ref{sec:method}. Prototype recovery builds a forget-class centroid
from $k{=}5$ shots in penultimate space and classifies held-out forget images by
nearest centroid (no weight update), averaged over five attack seeds. The linear
probe fits logistic regression on avgpool features across all classes, a
lower-variance summary independent of the centroid. Per-layer linear CKA is computed
between each unlearned model and its base on $n{=}200$ forget images. Head-only
relearning re-fits only the final linear layer on a held-out retain subset (no forget
labels, so $0\%$ recall is expected). Membership inference is a $10^3$-permutation
test used only as an output-level sanity check.

\paragraph{A.4\quad Equivalence testing and power} We assess ``the mask does nothing''
with the two one-sided tests procedure (TOST) \citep{schuirmann1987tost,%
lakens2017equivalence} at the pre-registered margin $\delta{=}\pm5$pp, fixed
\emph{a priori} at the level of the inter-seed SD of recovery ($\approx3$--$4$pp on
CIFAR-10), so a difference inside the margin is indistinguishable from seed noise.
Each paired observation is one class, seed-averaged over three seeds, so $n{=}10$. A
TOST $p<0.05$ certifies equivalence at the margin; we report the paired $t$-test $p$
alongside so magnitude and equivalence can be read together. On CIFAR-100 the few-shot
prototype metric carries per-class SD $\approx13$pp; a standard TOST power calculation
($\alpha{=}0.05$, power $0.8$) then requires $n\approx40$ classes to resolve the
comparison, four times our census, which is why the lower-variance linear probe is the
decisive metric there (\S\ref{sec:cifar100}).

\paragraph{A.5\quad Compute, software, and availability} The pipeline is implemented in
PyTorch with torchvision models and data, scikit-learn for the probe, silhouette, and
t-SNE, and SciPy for the statistical tests. Base and Gold training use a single CUDA
GPU on the DIBRIS laboratory server; the mask-geometry and mask-specificity
diagnostics (\S\ref{sec:mechanism}--\S\ref{sec:specificity}) are pure analysis on a
cached checkpoint and run on CPU in minutes. Checkpoints are cached per class so every
derived figure and table can be regenerated without repeating the $182$-epoch base
training. A self-contained audit script recomputes every reported value from the
bundled result files. Code and result data are available at the repository given in
the Data Availability statement. Table~\ref{tab:hyperparams} in
\ref{app:tables} collects every hyperparameter in one place.

\section{Data Behind the Main-Text Claims}
\label{app:data}

The main text reports seed-averaged, class-averaged summaries
(Tables~\ref{tab:rec}--\ref{tab:specificity}, Figures~\ref{fig:cka}--\ref{fig:gradenergy}).
This section gives the complete data those averages are built from, split by what
each half of the diagnosis needs: per-class outcomes for RQ1, and the mask's own
geometry, independent of any downstream outcome, for RQ2--RQ3.

\paragraph{Per-class outcomes (RQ1)} Table~\ref{tab:appendix_recovery} and
Table~\ref{tab:appendix_recovery_c100} in \ref{app:tables} give every number behind
Tables~\ref{tab:rec}--\ref{tab:specificity}, with no averaging of their own: all ten
classes, all three seeds, both datasets, thirty rows per table. Reading down the
recovery columns shows directly what \S\ref{sec:cifar100} states in aggregate, that
CIFAR-10 seeds agree to within a few points per class while CIFAR-100 seeds swing by
$20$--$44$pp on the same class (e.g.\ \emph{maple\_tree} under SalUn;
\emph{pear} and \emph{cloud} under RL-full);
reading across shows that the late-layer gradient share, the SalUn/Gold CKA
gap, and the near-$0.333$ specificity Jaccard already visible in
Figures~\ref{fig:gradenergy}--\ref{fig:cka} and Table~\ref{tab:specificity} hold on
every individual run, not only on average. Table~\ref{tab:appendix_robustness} there
reports the stability of the internal genuine-forgetting gate used as an
output-level sanity check throughout (\S\ref{sec:method}).

\paragraph{Mask geometry (RQ2--RQ3)} Two further checks isolate the mask itself,
independent of any recovery outcome. Table~\ref{tab:maskgeom}(a) in \ref{app:tables}
reports, for the realized SalUn mask at the default sparsity, what fraction
of its weight budget sits in each network region: layer~3, layer~4, and the head
together account for ${\approx}92\%$ of the mask on CIFAR-10, matching the
gradient-energy share behind Figure~\ref{fig:gradenergy} almost exactly, and
${\approx}91\%$ on CIFAR-100, somewhat more concentrated than the gradient's
${\approx}81\%$ because the mask applies one global magnitude threshold rather than
preserving the energy share exactly. This confirms directly, on the mask rather than
the gradient, that every condition is confined to the same late subspace before any
unlearning step runs. Table~\ref{tab:maskgeom}(b) complements the sparsity sweep of
\S\ref{sec:random} (three replicated classes there): here, over
the full ten-class census at every swept ratio, the raw Jaccard overlap between
SalUn's mask and an independently-drawn random mask of equal sparsity
matches the analytic chance-level null $r/(2-r)$ to within $0.0005$, on both datasets,
at every ratio. The saliency mask is therefore not only equivalent in outcome
(\S\ref{sec:random}) and weakly class-specific (\S\ref{sec:specificity}); at the level
of raw weight identity it is statistically indistinguishable from an independently
drawn random mask, at every sparsity we tested.

\section{Supplementary Tables}
\label{app:tables}

Every table named in \ref{app:repro} and \ref{app:data} is collected below, together
with two further tables that give the per-class data behind two claims the main text
states only in prose.

\paragraph{Cluster coherence (Table~\ref{tab:silhouette})} \S\ref{sec:mechanism}
reports that the forget-class silhouette stays ``in the same range as the original
trained model (${\approx}0.6$--$0.7$)'' under every mask, and collapses only under
Gold. Table~\ref{tab:silhouette} gives the value behind that range for all ten
CIFAR-10 classes at the primary seed, plus two replicate seeds for three of them,
so the ${\approx}0.6$--$0.7$ claim can be checked against every individual class and
seed rather than an average that could hide a class-specific exception.

\paragraph{Error redirection (Table~\ref{tab:redirection})} \S\ref{sec:random}
reports that on CIFAR-10 ``every one of the ten classes resolves to the same
consensus wrong class under SalUn, RL-full, and the random mask alike,''
and that on CIFAR-100 ``the three conditions agree on the dominant wrong class for
most classes and disagree only where a second candidate is close behind.''
Table~\ref{tab:redirection} gives the full per-class top-1 wrong-label census this
claim is drawn from, on both datasets at the primary seed.

\begin{table*}[p]
\centering
\scriptsize\setlength{\tabcolsep}{5pt}
\renewcommand{\arraystretch}{0.85}
\caption{Full experimental configuration (\S A.1--A.4). Every value is held identical
across the three unlearning conditions and both datasets; only the mask $\mathbf{m}$
(Eq.~\ref{eq:mask}) changes between conditions.}
\label{tab:hyperparams}
\makeatletter
\if@twocolumn
\begin{minipage}[t]{0.49\linewidth}
\begin{tabular}{@{}ll@{}}
\toprule
Setting & Value \\
\midrule
\multicolumn{2}{@{}l}{\textit{Architecture \& data}}\\
Backbone & ResNet-18 \citep{he2016resnet} \\
Penultimate dim.\ & $512$ \\
Datasets & CIFAR-10, CIFAR-100 \citep{krizhevsky2009cifar} \\
Input & $32\times32$ RGB \\
Augmentation & RandomCrop($32$,pad $4$), HFlip, \\
 & AutoAugment (C10), RandomErasing ($p{=}0.2$) \\
\addlinespace[1pt]
\multicolumn{2}{@{}l}{\textit{Base model (original)}}\\
Optimizer & AdamW, wd $5\times10^{-4}$ \\
Learning rate & $1\times10^{-3}$, cosine, $5$-epoch warmup \\
Epochs / batch & $182$ / $128$, mixed precision \\
\addlinespace[1pt]
\multicolumn{2}{@{}l}{\textit{Gold retrain (exact unlearning)}}\\
Data & retain set $\mathcal{D}_r$ only \\
Optimizer & AdamW, lr $1\times10^{-3}$, cosine \\
Epochs & $40$, $3$-epoch warmup \\
\bottomrule
\end{tabular}
\end{minipage}%
\hfill
\begin{minipage}[t]{0.49\linewidth}
\begin{tabular}{@{}ll@{}}
\toprule
Setting & Value \\
\midrule
\multicolumn{2}{@{}l}{\textit{Unlearning update (all three conditions)}}\\
Objective & random-label RL + retain CE (Eq.~\ref{eq:rl}) \\
Optimizer & SGD, momentum $0.9$, wd $5\times10^{-4}$ \\
Learning rate & $0.013$ (SalUn default) \\
Repair epochs & $10$ \\
Mask sparsity & $0.5$; swept $\{0.1,0.3,0.5,0.7\}$ \\
Forget set & one class ($5000$ / $500$ imgs, C10 / C100) \\
\addlinespace[1pt]
\multicolumn{2}{@{}l}{\textit{Evaluation}}\\
Prototype recovery & $k{=}5$, mean over $5$ attack seeds \\
Linear probe & logistic reg.\ on avgpool features \\
Per-layer CKA & linear CKA \citep{kornblith2019cka}, $n{=}200$ \\
Membership inference & permutation test, $10^3$ perms \\
Equivalence test & paired TOST, $\delta{=}\pm5$pp, $\alpha{=}0.05$ \\
\addlinespace[1pt]
\multicolumn{2}{@{}l}{\textit{Protocol}}\\
Seeds & $42$ (primary), $1337$, $999$ \\
Runs & $10$ classes $\times$ $3$ seeds $\times$ $3$ cond.\ per dataset \\
\bottomrule
\end{tabular}
\end{minipage}
\else
\centering
\resizebox{0.88\linewidth}{!}{%
\begin{tabular}{@{}ll@{}}
\toprule
Setting & Value \\
\midrule
\multicolumn{2}{@{}l}{\textit{Architecture \& data}}\\
Backbone & ResNet-18 \citep{he2016resnet} \\
Penultimate dim.\ & $512$ \\
Datasets & CIFAR-10, CIFAR-100 \citep{krizhevsky2009cifar} \\
Input & $32\times32$ RGB \\
Augmentation & RandomCrop($32$,pad $4$), HFlip, AutoAugment (C10), RandomErasing ($p{=}0.2$) \\
\addlinespace[1pt]
\multicolumn{2}{@{}l}{\textit{Base model (original)}}\\
Optimizer & AdamW, wd $5\times10^{-4}$ \\
Learning rate & $1\times10^{-3}$, cosine, $5$-epoch warmup \\
Epochs / batch & $182$ / $128$, mixed precision \\
\addlinespace[1pt]
\multicolumn{2}{@{}l}{\textit{Gold retrain (exact unlearning)}}\\
Data & retain set $\mathcal{D}_r$ only \\
Optimizer & AdamW, lr $1\times10^{-3}$, cosine \\
Epochs & $40$, $3$-epoch warmup \\
\addlinespace[1pt]
\multicolumn{2}{@{}l}{\textit{Unlearning update (all three conditions)}}\\
Objective & random-label RL + retain CE (Eq.~\ref{eq:rl}) \\
Optimizer & SGD, momentum $0.9$, wd $5\times10^{-4}$ \\
Learning rate & $0.013$ (SalUn default) \\
Repair epochs & $10$ \\
Mask sparsity & $0.5$; swept $\{0.1,0.3,0.5,0.7\}$ \\
Forget set & one class ($5000$ / $500$ imgs, C10 / C100) \\
\addlinespace[1pt]
\multicolumn{2}{@{}l}{\textit{Evaluation}}\\
Prototype recovery & $k{=}5$, mean over $5$ attack seeds \\
Linear probe & logistic reg.\ on avgpool features \\
Per-layer CKA & linear CKA \citep{kornblith2019cka}, $n{=}200$ \\
Membership inference & permutation test, $10^3$ perms \\
Equivalence test & paired TOST, $\delta{=}\pm5$pp, $\alpha{=}0.05$ \\
\addlinespace[1pt]
\multicolumn{2}{@{}l}{\textit{Protocol}}\\
Seeds & $42$ (primary), $1337$, $999$ \\
Runs & $10$ classes $\times$ $3$ seeds $\times$ $3$ cond.\ per dataset \\
\bottomrule
\end{tabular}}
\fi
\makeatother
\end{table*}

\begin{table*}[p]\centering
\scriptsize\setlength{\tabcolsep}{3pt}
\renewcommand{\arraystretch}{0.70}
\caption{Complete per-class, per-seed data behind every average in Tables~\ref{tab:rec}--\ref{tab:specificity}, Table~\ref{tab:causal}, and Figures~\ref{fig:perclass}--\ref{fig:gradenergy} (see \ref{app:data} for how to read this table): CIFAR-10, all $10$ classes $\times$ $3$ seeds, no averaging.}
\label{tab:appendix_recovery}
\resizebox{0.92\linewidth}{!}{%
\begin{tabular}{@{}llrrrrrrrrrrrr@{}}\toprule
& & \multicolumn{7}{c}{Recovery (\%)} & Grad.\ & \multicolumn{2}{c}{CKA-L4} & \multicolumn{2}{c}{Specificity $J$} \\
\cmidrule(lr){3-9}\cmidrule(lr){11-12}\cmidrule(lr){13-14}
Class & Seed & \multicolumn{1}{c}{SalUn} & \multicolumn{1}{c}{Random} & \multicolumn{1}{c}{RL-full} & \multicolumn{1}{c}{Fisher} & \multicolumn{1}{c}{Early} & \multicolumn{1}{c}{E-$3\times$} & \multicolumn{1}{c}{Gold} & \multicolumn{1}{c}{Late-\%} & \multicolumn{1}{c}{SalUn} & \multicolumn{1}{c}{Gold} & \multicolumn{1}{c}{Other} & \multicolumn{1}{c}{Retain} \\
\midrule
\addlinespace[1pt]
airplane & 42 & 92.0 & 95.3 & 96.5 & 97.0 & 93.0 & 92.5 & 38.6 & 98.7 & 0.571 & 0.209 & 0.391 & 0.398 \\
 & 1337 & 96.4 & 96.6 & 95.8 & 95.0 & 95.1 & 94.0 & 38.5 & 83.5 & 0.624 & 0.209 & 0.410 & 0.377 \\
 & 999 & 94.2 & 95.0 & 95.6 & 97.2 & 95.1 & 92.1 & 38.7 & 89.4 & 0.596 & 0.209 & 0.401 & 0.383 \\
\addlinespace[0.8pt]
automobile & 42 & 98.0 & 97.1 & 98.4 & 96.0 & 98.2 & 98.0 & 34.1 & 99.8 & 0.569 & 0.102 & 0.403 & 0.430 \\
 & 1337 & 98.0 & 97.2 & 97.3 & 97.4 & 96.3 & 94.7 & 33.8 & 96.9 & 0.587 & 0.102 & 0.395 & 0.381 \\
 & 999 & 96.6 & 97.7 & 98.8 & 97.4 & 97.9 & 96.2 & 33.2 & 98.2 & 0.555 & 0.102 & 0.385 & 0.375 \\
\addlinespace[0.8pt]
bird & 42 & 93.1 & 93.7 & 91.6 & 92.3 & 87.2 & 86.9 & 39.9 & 99.6 & 0.819 & 0.110 & 0.396 & 0.426 \\
 & 1337 & 88.2 & 93.4 & 89.2 & 91.4 & 85.3 & 82.6 & 39.7 & 85.7 & 0.756 & 0.110 & 0.411 & 0.393 \\
 & 999 & 93.5 & 93.0 & 94.4 & 89.7 & 88.8 & 85.2 & 39.6 & 89.7 & 0.808 & 0.110 & 0.401 & 0.395 \\
\addlinespace[0.8pt]
cat & 42 & 89.0 & 85.6 & 90.3 & 86.5 & 87.3 & 87.5 & 37.1 & 99.5 & 0.687 & 0.143 & 0.398 & 0.411 \\
 & 1337 & 81.0 & 83.1 & 83.7 & 83.6 & 80.0 & 83.3 & 37.0 & 73.2 & 0.625 & 0.143 & 0.436 & 0.403 \\
 & 999 & 84.1 & 83.0 & 84.1 & 84.9 & 80.9 & 86.4 & 37.1 & 77.5 & 0.616 & 0.143 & 0.424 & 0.382 \\
\addlinespace[0.8pt]
deer & 42 & 92.3 & 97.1 & 95.8 & 94.2 & 88.1 & 94.1 & 45.5 & 99.5 & 0.757 & 0.120 & 0.394 & 0.415 \\
 & 1337 & 91.4 & 94.7 & 94.2 & 96.3 & 89.5 & 94.4 & 45.4 & 91.0 & 0.702 & 0.120 & 0.397 & 0.395 \\
 & 999 & 93.0 & 94.1 & 94.2 & 93.7 & 89.5 & 93.3 & 45.6 & 91.5 & 0.647 & 0.120 & 0.384 & 0.381 \\
\addlinespace[0.8pt]
dog & 42 & 91.2 & 88.9 & 89.5 & 91.5 & 84.3 & 80.4 & 37.1 & 98.7 & 0.620 & 0.061 & 0.376 & 0.402 \\
 & 1337 & 92.4 & 89.3 & 87.9 & 87.5 & 85.8 & 82.6 & 37.1 & 73.6 & 0.545 & 0.061 & 0.384 & 0.403 \\
 & 999 & 94.1 & 89.6 & 80.8 & 92.6 & 82.8 & 73.7 & 36.8 & 76.2 & 0.517 & 0.061 & 0.380 & 0.388 \\
\addlinespace[0.8pt]
frog & 42 & 89.2 & 94.9 & 94.1 & 96.8 & 91.6 & 92.0 & 48.6 & 99.6 & 0.676 & 0.094 & 0.399 & 0.418 \\
 & 1337 & 97.4 & 95.9 & 96.5 & 94.2 & 94.7 & 90.2 & 48.5 & 86.8 & 0.567 & 0.094 & 0.411 & 0.392 \\
 & 999 & 89.9 & 94.3 & 95.3 & 96.7 & 97.7 & 77.8 & 48.8 & 85.9 & 0.649 & 0.094 & 0.403 & 0.389 \\
\addlinespace[0.8pt]
horse & 42 & 96.4 & 97.4 & 93.9 & 88.8 & 97.3 & 94.4 & 44.6 & 99.8 & 0.640 & 0.182 & 0.396 & 0.419 \\
 & 1337 & 93.1 & 97.0 & 92.8 & 96.0 & 92.0 & 93.2 & 44.5 & 92.0 & 0.619 & 0.182 & 0.400 & 0.381 \\
 & 999 & 96.6 & 94.9 & 96.7 & 90.1 & 97.1 & 95.2 & 44.6 & 91.1 & 0.583 & 0.182 & 0.392 & 0.367 \\
\addlinespace[0.8pt]
ship & 42 & 94.5 & 95.8 & 96.6 & 96.1 & 97.3 & 96.2 & 38.8 & 99.7 & 0.652 & 0.090 & 0.390 & 0.422 \\
 & 1337 & 96.0 & 96.0 & 95.9 & 94.5 & 97.0 & 93.5 & 38.5 & 94.4 & 0.583 & 0.090 & 0.394 & 0.379 \\
 & 999 & 97.4 & 96.3 & 97.1 & 95.9 & 97.1 & 94.9 & 38.9 & 92.8 & 0.426 & 0.090 & 0.383 & 0.376 \\
\addlinespace[0.8pt]
truck & 42 & 94.5 & 97.8 & 97.4 & 96.6 & 98.8 & 96.7 & 51.8 & 99.8 & 0.453 & 0.103 & 0.400 & 0.411 \\
 & 1337 & 97.0 & 96.2 & 95.7 & 96.5 & 97.7 & 95.6 & 51.8 & 95.9 & 0.458 & 0.103 & 0.397 & 0.385 \\
 & 999 & 96.6 & 94.8 & 97.7 & 97.0 & 99.2 & 95.1 & 51.7 & 95.6 & 0.612 & 0.103 & 0.390 & 0.379 \\
\bottomrule
\end{tabular}}

\smallskip
{\scriptsize Fisher/Early/E-$3\times$: the second selection criterion and the causal placement conditions of \S\ref{sec:causal}. CKA-L4: SalUn/Gold vs.\ original, layer 4. Other/Retain $J$: mask Jaccard vs.\ other-class / retain-set gradients (null $\approx0.333$).\par}
\end{table*}

\begin{table*}[p]\centering
\footnotesize\setlength{\tabcolsep}{5pt}
\caption{Robustness of the genuine-forgetting decision rule used as an output-level sanity check: a run is flagged as genuine forgetting only if all three Gold-normalized residual scores (noise-corrected membership inference, probe-recall excess over Gold, and penultimate-CKA excess over Gold, each clipped to $[0,1]$) fall below $\tau{=}0.20$. Across the 30 class-seed combinations per dataset: fraction where the $\tau{=}0.20$ threshold separates conditions cleanly, where the flag's bootstrap CI is stable ($B{=}500$), and where the flag is unchanged under $\pm0.05$ noise-floor perturbation.}
\label{tab:appendix_robustness}
\resizebox{0.55\linewidth}{!}{%
\begin{tabular}{@{}lccc@{}}\toprule
Dataset & $\tau{=}0.20$ usable & Bootstrap stable & Noise-floor robust \\
\midrule
CIFAR-10 & 27/30 \;(90\%) & 149/150 \;(99\%) & 90/90 \;(100\%) \\
CIFAR-100 & 27/30 \;(90\%) & 52/150 \;(35\%) & 67/90 \;(74\%) \\
\bottomrule
\end{tabular}}

\smallskip
{\footnotesize The CIFAR-100 bootstrap instability (52/150) mirrors the high per-class variance of the few-shot prototype metric already discussed in \S\ref{sec:cifar100}: Gold's own CI is unstable just as often as the three unlearning conditions', so this is a property of the dataset's fine-grained geometry, not a mask-specific weakness.\par}
\end{table*}

\begin{table*}[p]\centering
\scriptsize\setlength{\tabcolsep}{3pt}
\renewcommand{\arraystretch}{0.70}
\caption{Same as Table~\ref{tab:appendix_recovery}, for CIFAR-100.}
\label{tab:appendix_recovery_c100}
\resizebox{0.92\linewidth}{!}{%
\begin{tabular}{@{}llrrrrrrrrrrrr@{}}\toprule
& & \multicolumn{7}{c}{Recovery (\%)} & Grad.\ & \multicolumn{2}{c}{CKA-L4} & \multicolumn{2}{c}{Specificity $J$} \\
\cmidrule(lr){3-9}\cmidrule(lr){11-12}\cmidrule(lr){13-14}
Class & Seed & \multicolumn{1}{c}{SalUn} & \multicolumn{1}{c}{Random} & \multicolumn{1}{c}{RL-full} & \multicolumn{1}{c}{Fisher} & \multicolumn{1}{c}{Early} & \multicolumn{1}{c}{E-$3\times$} & \multicolumn{1}{c}{Gold} & \multicolumn{1}{c}{Late-\%} & \multicolumn{1}{c}{SalUn} & \multicolumn{1}{c}{Gold} & \multicolumn{1}{c}{Other} & \multicolumn{1}{c}{Retain} \\
\midrule
\addlinespace[1pt]
apple & 42 & 53.1 & 58.1 & 56.0 & 65.3 & 78.7 & 72.6 & 59.4 & 90.4 & 0.457 & 0.627 & 0.408 & 0.405 \\
 & 1337 & 38.7 & 58.9 & 52.8 & 44.6 & 82.3 & 70.5 & 59.4 & 82.3 & 0.318 & 0.627 & 0.406 & 0.403 \\
 & 999 & 59.6 & 41.5 & 64.6 & 60.0 & 82.1 & 79.6 & 58.9 & 83.1 & 0.275 & 0.627 & 0.401 & 0.398 \\
\addlinespace[0.8pt]
bus & 42 & 6.5 & 11.8 & 14.3 & 11.2 & 31.8 & 30.3 & 15.6 & 90.0 & 0.596 & 0.662 & 0.410 & 0.398 \\
 & 1337 & 6.9 & 7.6 & 7.6 & 17.1 & 31.4 & 40.8 & 16.2 & 84.1 & 0.563 & 0.662 & 0.404 & 0.407 \\
 & 999 & 9.5 & 18.9 & 12.0 & 13.1 & 26.7 & 34.3 & 16.2 & 84.5 & 0.541 & 0.662 & 0.404 & 0.409 \\
\addlinespace[0.8pt]
cloud & 42 & 22.9 & 16.8 & 12.0 & 21.3 & 57.3 & 63.6 & 25.5 & 90.4 & 0.368 & 0.479 & 0.393 & 0.391 \\
 & 1337 & 38.1 & 33.3 & 23.2 & 32.8 & 74.7 & 58.7 & 25.7 & 73.9 & 0.437 & 0.479 & 0.393 & 0.379 \\
 & 999 & 39.2 & 26.1 & 53.9 & 38.5 & 65.3 & 53.9 & 25.5 & 66.9 & 0.412 & 0.479 & 0.398 & 0.386 \\
\addlinespace[0.8pt]
dolphin & 42 & 27.8 & 12.0 & 14.3 & 17.1 & 62.3 & 57.9 & 17.7 & 86.1 & 0.472 & 0.546 & 0.411 & 0.389 \\
 & 1337 & 25.5 & 15.4 & 13.1 & 21.3 & 66.1 & 52.4 & 18.1 & 78.3 & 0.387 & 0.546 & 0.410 & 0.394 \\
 & 999 & 33.7 & 27.2 & 19.4 & 27.6 & 66.3 & 55.2 & 17.9 & 73.0 & 0.367 & 0.546 & 0.407 & 0.394 \\
\addlinespace[0.8pt]
lion & 42 & 29.7 & 54.5 & 49.9 & 16.6 & 78.1 & 69.5 & 28.0 & 84.5 & 0.387 & 0.438 & 0.417 & 0.410 \\
 & 1337 & 17.9 & 17.7 & 32.2 & 35.8 & 63.4 & 69.7 & 28.4 & 73.7 & 0.439 & 0.438 & 0.410 & 0.414 \\
 & 999 & 29.1 & 25.3 & 53.9 & 10.7 & 69.7 & 73.7 & 28.2 & 76.8 & 0.537 & 0.438 & 0.410 & 0.413 \\
\addlinespace[0.8pt]
maple\_tree & 42 & 20.6 & 12.8 & 26.7 & 22.5 & 38.3 & 37.9 & 15.4 & 75.2 & 0.530 & 0.541 & 0.411 & 0.391 \\
 & 1337 & 14.5 & 16.6 & 56.8 & 20.0 & 26.7 & 50.1 & 14.9 & 60.0 & 0.454 & 0.541 & 0.411 & 0.397 \\
 & 999 & 42.3 & 18.5 & 12.8 & 10.1 & 44.2 & 50.9 & 15.4 & 66.0 & 0.302 & 0.541 & 0.409 & 0.390 \\
\addlinespace[0.8pt]
pear & 42 & 15.2 & 19.6 & 3.8 & 20.4 & 41.7 & 38.7 & 21.5 & 91.9 & 0.383 & 0.464 & 0.399 & 0.404 \\
 & 1337 & 10.7 & 15.4 & 5.1 & 15.6 & 41.5 & 43.6 & 21.7 & 79.3 & 0.372 & 0.464 & 0.403 & 0.397 \\
 & 999 & 14.5 & 14.7 & 48.2 & 12.8 & 47.6 & 28.6 & 21.7 & 79.6 & 0.409 & 0.464 & 0.401 & 0.391 \\
\addlinespace[0.8pt]
rose & 42 & 3.6 & 11.6 & 24.6 & 10.3 & 60.6 & 77.3 & 9.7 & 90.8 & 0.508 & 0.524 & 0.406 & 0.396 \\
 & 1337 & 13.7 & 17.1 & 17.5 & 24.8 & 58.7 & 54.3 & 10.1 & 79.3 & 0.527 & 0.524 & 0.412 & 0.402 \\
 & 999 & 20.8 & 13.5 & 32.0 & 24.2 & 61.1 & 78.9 & 9.5 & 79.0 & 0.540 & 0.524 & 0.413 & 0.404 \\
\addlinespace[0.8pt]
tank & 42 & 17.1 & 54.5 & 18.9 & 11.6 & 67.4 & 67.4 & 18.9 & 85.0 & 0.314 & 0.433 & 0.396 & 0.411 \\
 & 1337 & 17.3 & 13.1 & 29.3 & 20.2 & 68.4 & 79.8 & 18.7 & 80.3 & 0.386 & 0.433 & 0.395 & 0.403 \\
 & 999 & 36.4 & 29.9 & 14.1 & 39.8 & 70.5 & 78.3 & 18.5 & 79.9 & 0.280 & 0.433 & 0.399 & 0.403 \\
\addlinespace[0.8pt]
bicycle & 42 & 33.9 & 68.6 & 30.7 & 32.2 & 67.4 & 69.7 & 42.1 & 90.9 & 0.380 & 0.508 & 0.404 & 0.408 \\
 & 1337 & 34.7 & 33.3 & 10.7 & 39.2 & 66.7 & 59.6 & 41.7 & 84.4 & 0.386 & 0.508 & 0.400 & 0.411 \\
 & 999 & 32.8 & 36.4 & 11.2 & 31.2 & 73.1 & 84.8 & 41.9 & 81.5 & 0.383 & 0.508 & 0.400 & 0.407 \\
\bottomrule
\end{tabular}}

\smallskip
{\scriptsize Column definitions as in Table~\ref{tab:appendix_recovery}.\par}
\end{table*}

\begin{table*}[p]\centering
\scriptsize\setlength{\tabcolsep}{4pt}
\renewcommand{\arraystretch}{0.78}
\caption{\textbf{Mask geometry (\ref{app:data}): receipts for \S\ref{sec:mechanism}
and \S\ref{sec:specificity}, measured on the mask itself, not on outcomes.}
(a)~Depth profile of the default-sparsity SalUn mask. (b)~Saliency-mask vs.\
random-mask Jaccard overlap, excess above the analytic null $r/(2-r)$ (units of
$10^{-3}$). (c)~Saliency-mask vs.\ diagonal-Fisher-mask overlap: the second selection
criterion picks nearly the same weights. All panels: $10$ classes $\times$ $3$ seeds
per dataset.}
\label{tab:maskgeom}

\textbf{(a) Mask weight-budget share by network region.}\par
\resizebox{0.88\linewidth}{!}{%
\begin{tabular}{@{}lrrrr@{}}
\toprule
& \multicolumn{2}{c}{CIFAR-10} & \multicolumn{2}{c}{CIFAR-100} \\
\cmidrule(lr){2-3}\cmidrule(lr){4-5}
Region & Share (\%) & Region kept (\%) & Share (\%) & Region kept (\%) \\
\midrule
stem   & 0.03 \,[0.03, 0.03]   & 92.11 & 0.03 \,[0.03, 0.03]   & 96.23 \\
layer1 & 1.79 \,[1.0, 2.2]     & 67.48 & 2.10 \,[1.9, 2.4]     & 79.52 \\
layer2 & 6.09 \,[3.4, 7.8]     & 64.69 & 7.14 \,[6.6, 7.9]     & 76.25 \\
layer3 & 20.54 \,[11.7, 28.9]  & 54.65 & 24.36 \,[22.3, 27.1]  & 65.08 \\
layer4 & 71.47 \,[61.0, 83.2]  & 47.57 & 66.12 \,[62.7, 68.7]  & 44.19 \\
head   & 0.09 \,[0.09, 0.09]   & 99.18 & 0.25 \,[0.11, 0.44]   & 27.75 \\
\bottomrule
\end{tabular}}

\medskip
\textbf{(b) Saliency mask vs.\ an independent random mask of equal sparsity.}\par
\resizebox{0.88\linewidth}{!}{%
\begin{tabular}{@{}lrrrr@{}}
\toprule
& \multicolumn{2}{c}{CIFAR-10} & \multicolumn{2}{c}{CIFAR-100} \\
\cmidrule(lr){2-3}\cmidrule(lr){4-5}
Sparsity $r$ & Mean excess & Max $|$excess$|$ & Mean excess & Max $|$excess$|$ \\
\midrule
0.10 & $+0.01$ & 0.3 & $+0.05$ & 0.3 \\
0.30 & $+0.02$ & 0.3 & $+0.03$ & 0.4 \\
0.50 & $-0.07$ & 0.5 & $-0.01$ & 0.3 \\
0.70 & $+0.00$ & 0.2 & $-0.03$ & 0.3 \\
\bottomrule
\end{tabular}}

\medskip
\textbf{(c) Saliency mask vs.\ the diagonal-Fisher mask at equal sparsity ($r{=}0.5$).}\par
\resizebox{0.6\linewidth}{!}{%
\begin{tabular}{@{}lrr@{}}
\toprule
& $J$(Fisher, SalUn) & Fisher late share (\%) \\
\midrule
CIFAR-10  & $0.832$ & $89.4$ \\
CIFAR-100 & $0.937$ & $89.8$ \\
\bottomrule
\end{tabular}}

\smallskip
{\scriptsize (a) Layer3+layer4+head sums to $92.10\%$ (C10) / $90.73\%$ (C100); region-kept does not sum to $100\%$. (b) excess $=$ Jaccard $-\,r/(2-r)$, one random draw per class-seed-ratio, $30$ draws/cell. (c) $10$ classes $\times$ $3$ seeds per dataset; null $= 1/3$: the two selection criteria pick nearly the same weights, and both place ${\approx}89\%$ of their budget in layer3+layer4+head.\par}
\end{table*}

\begin{table*}[p]\centering
\scriptsize\setlength{\tabcolsep}{4pt}
\renewcommand{\arraystretch}{0.78}
\caption{\textbf{The forget class stays a coherent cluster under every mask.} Cosine
silhouette of the forget class in raw 512-d penultimate features, CIFAR-10, primary
seed ($42$) for all ten classes plus two replicate seeds for three of them (airplane,
automobile, cat). SalUn, RL-full, Random Mask, and the Fisher mask all stay
near the base model's silhouette; the early mask (and its forced variant) sits visibly
lower, distorting the cluster's geometry more, yet remains far above Gold, which alone
collapses it toward (and below) $0$.}
\label{tab:silhouette}
\resizebox{\linewidth}{!}{%
\begin{tabular}{@{}llrrrrrrrr@{}}\toprule
Class & Seed & Base & Gold & SalUn-RL & RL-full & Random Mask & Fisher & Early & Early $3\times$ \\
\midrule
airplane & 42 & 0.644 & $-0.114$ & 0.541 & 0.679 & 0.689 & 0.651 & 0.458 & 0.479 \\
 & 1337 & 0.644 & $-0.114$ & 0.684 & 0.711 & 0.738 & 0.628 & 0.481 & 0.535 \\
 & 999 & 0.644 & $-0.114$ & 0.602 & 0.630 & 0.610 & 0.670 & 0.514 & 0.445 \\
\addlinespace[0.8pt]
automobile & 42 & 0.749 & $-0.078$ & 0.722 & 0.782 & 0.709 & 0.648 & 0.756 & 0.772 \\
 & 1337 & 0.749 & $-0.078$ & 0.734 & 0.758 & 0.718 & 0.740 & 0.628 & 0.661 \\
 & 999 & 0.749 & $-0.078$ & 0.654 & 0.775 & 0.700 & 0.644 & 0.718 & 0.736 \\
\addlinespace[0.8pt]
bird & 42 & 0.605 & $-0.153$ & 0.599 & 0.571 & 0.620 & 0.525 & 0.317 & 0.341 \\
\addlinespace[0.8pt]
cat & 42 & 0.499 & $-0.161$ & 0.507 & 0.629 & 0.460 & 0.472 & 0.393 & 0.331 \\
 & 1337 & 0.499 & $-0.161$ & 0.320 & 0.442 & 0.370 & 0.398 & 0.323 & 0.292 \\
 & 999 & 0.499 & $-0.161$ & 0.400 & 0.426 & 0.401 & 0.464 & 0.295 & 0.342 \\
\addlinespace[0.8pt]
deer & 42 & 0.702 & $-0.038$ & 0.650 & 0.737 & 0.747 & 0.647 & 0.503 & 0.567 \\
\addlinespace[0.8pt]
dog & 42 & 0.546 & $-0.049$ & 0.639 & 0.624 & 0.601 & 0.594 & 0.401 & 0.400 \\
\addlinespace[0.8pt]
frog & 42 & 0.666 & 0.019 & 0.512 & 0.661 & 0.675 & 0.729 & 0.465 & 0.476 \\
\addlinespace[0.8pt]
horse & 42 & 0.707 & $-0.052$ & 0.674 & 0.675 & 0.764 & 0.503 & 0.748 & 0.551 \\
\addlinespace[0.8pt]
ship & 42 & 0.749 & $-0.032$ & 0.680 & 0.693 & 0.690 & 0.624 & 0.682 & 0.702 \\
\addlinespace[0.8pt]
truck & 42 & 0.724 & $-0.195$ & 0.630 & 0.751 & 0.764 & 0.680 & 0.835 & 0.679 \\
\bottomrule
\end{tabular}}

\smallskip
{\scriptsize Base and Gold are a single per-class model (not reseeded), so their values repeat
across the extra-seed rows; only the unlearning conditions vary with seed. Ten-class means at
the primary seed: Base $0.66$, Gold $-0.09$, SalUn $0.62$, RL-full $0.68$,
Random $0.67$, Fisher $0.61$, Early $0.56$, Early~$3\times$ $0.53$.\par}
\end{table*}

\begin{table*}[p]\centering
\scriptsize\setlength{\tabcolsep}{4pt}
\renewcommand{\arraystretch}{0.82}
\caption{\textbf{Where does the forgotten class go?} Most common (top-1) wrong label
assigned to held-out forget-class images, primary seed, by condition; baseline is the
class's own correct-classification rate before unlearning. \textbf{Bold} wrong-class
names mark full three-way consensus.}
\label{tab:redirection}
\resizebox{0.88\linewidth}{!}{%
\begin{tabular}{@{}lrlll@{}}\toprule
Class & Baseline correct & SalUn & RL-full & Random Mask \\
\midrule
\multicolumn{5}{@{}l}{\textit{CIFAR-10}}\\
airplane & 0.95 & \textbf{automobile} (0.83) & \textbf{automobile} (0.91) & \textbf{automobile} (0.89) \\
automobile & 0.98 & \textbf{bird} (0.96) & \textbf{bird} (0.96) & \textbf{bird} (0.94) \\
bird & 0.93 & \textbf{cat} (0.88) & \textbf{cat} (0.65) & \textbf{cat} (0.84) \\
cat & 0.87 & \textbf{deer} (0.84) & \textbf{deer} (0.87) & \textbf{deer} (0.66) \\
deer & 0.95 & \textbf{dog} (0.87) & \textbf{dog} (0.93) & \textbf{dog} (0.93) \\
dog & 0.91 & \textbf{frog} (0.86) & \textbf{frog} (0.88) & \textbf{frog} (0.70) \\
frog & 0.97 & \textbf{horse} (0.81) & \textbf{horse} (0.90) & \textbf{horse} (0.88) \\
horse & 0.97 & \textbf{ship} (0.95) & \textbf{ship} (0.90) & \textbf{ship} (0.92) \\
ship & 0.97 & \textbf{truck} (0.93) & \textbf{truck} (0.96) & \textbf{truck} (0.94) \\
truck & 0.97 & \textbf{airplane} (0.93) & \textbf{airplane} (0.95) & \textbf{airplane} (0.94) \\
\addlinespace[1.5pt]
\multicolumn{5}{@{}l}{\textit{CIFAR-100}}\\
apple & 0.88 & \textbf{aquarium\_fish} (0.54) & \textbf{aquarium\_fish} (0.72) & \textbf{aquarium\_fish} (0.71) \\
bicycle & 0.93 & \textbf{bottle} (0.54) & \textbf{bottle} (0.70) & \textbf{bottle} (0.48) \\
bus & 0.69 & streetcar (0.37) & butterfly (0.39) & butterfly (0.24)$^\dagger$ \\
cloud & 0.87 & cockroach (0.50) & cockroach (0.64) & shrew (0.21) \\
dolphin & 0.71 & whale (0.26) & whale (0.31) & elephant (0.42) \\
lion & 0.85 & \textbf{lizard} (0.64) & \textbf{lizard} (0.74) & \textbf{lizard} (0.36) \\
maple\_tree & 0.67 & willow\_tree (0.30) & motorcycle (0.42) & oak\_tree (0.32)$^\dagger$ \\
pear & 0.80 & \textbf{pickup\_truck} (0.56) & \textbf{pickup\_truck} (0.67) & \textbf{pickup\_truck} (0.52) \\
rose & 0.82 & \textbf{sea} (0.31) & \textbf{sea} (0.42) & \textbf{sea} (0.32) \\
tank & 0.86 & \textbf{telephone} (0.70) & \textbf{telephone} (0.50) & \textbf{telephone} (0.56) \\
\bottomrule
\end{tabular}}

\smallskip
{\scriptsize All ten CIFAR-10 classes reach full three-way consensus; on CIFAR-100 six of ten do
(bus, cloud, dolphin, maple\_tree do not: the runner-up label is close enough behind that
different conditions land on different top-1 labels). $^\dagger$bus and maple\_tree, Random
Mask: 2-decimal tie in the source log (butterfly/streetcar 0.24/0.24; oak\_tree/motorcycle
0.32/0.32); the listed label is the first-parsed, not a resolved winner.\par}
\end{table*}

%% ---- Author biographies (typeset at the end, Elsevier style) ----
\clearpage
\newcommand{\authorbio}[3]{%
  \noindent
  \begin{minipage}[t]{1.0in}\vspace{0pt}\includegraphics[width=0.9in,keepaspectratio]{#1}\end{minipage}\hfill
  \begin{minipage}[t]{\dimexpr\linewidth-1.15in\relax}\vspace{0pt}{\bfseries #2} #3\end{minipage}%
  \par\medskip
}
\section*{Author biographies}

\authorbio{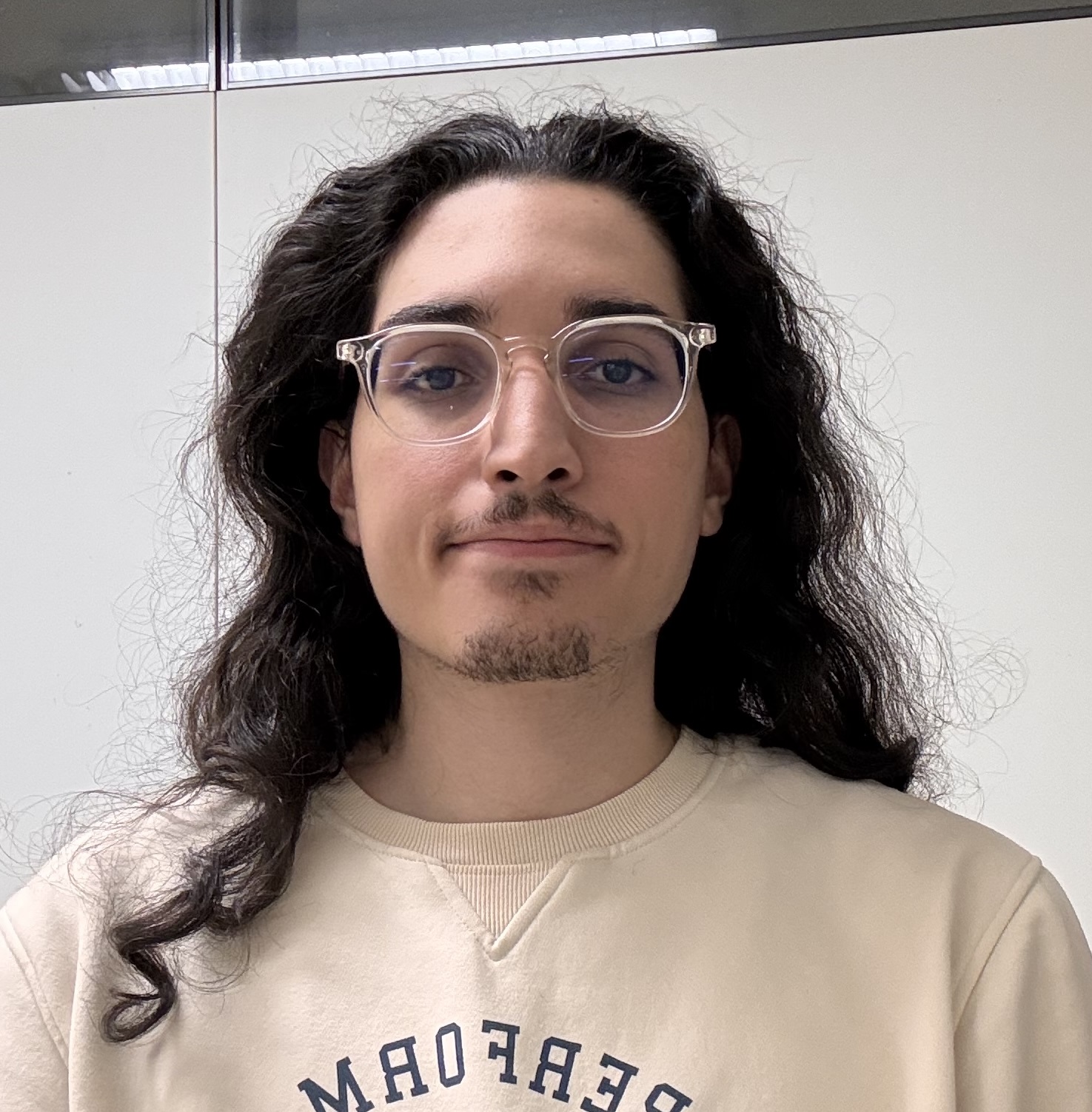}{Billel Habbati}{received the M.Sc. degree in Networks and Telecommunications from the University of Mentouri~1, Constantine, Algeria, in 2025. He is currently a Ph.D. student at the University of Genova, Italy. His research interests include machine unlearning, representation-level analysis of deep networks, and the security and verifiability of federated learning systems.}

\authorbio{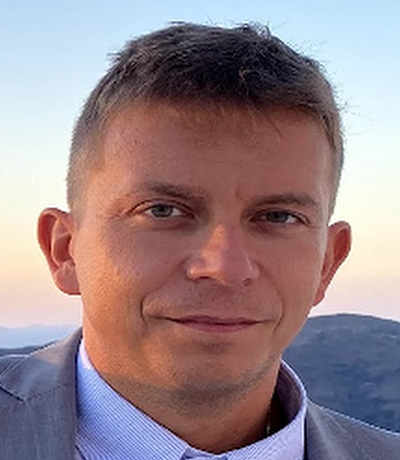}{Alessio Merlo}{received the Ph.D. degree in Computer Science from the University of Genova in 2010. He is currently a Professor in Computer Engineering and the Director of the Centre for Defense Higher Studies (CASD), Rome, Italy. He has published more than 120 scientific papers in international conferences and journals. His research interests include mobile security, the Internet of Things, and cyber-physical systems security.}

\authorbio{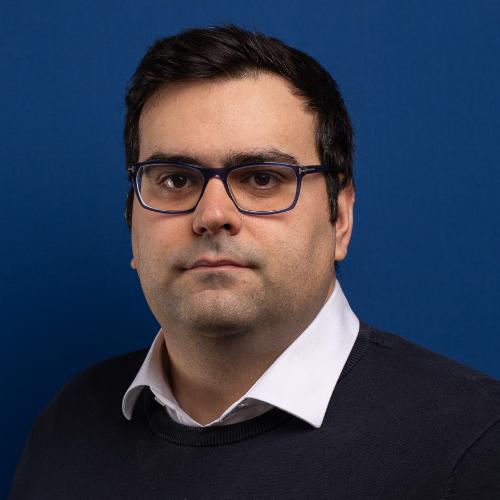}{Luca Verderame}{received the M.Sc. degree (cum laude) in 2011 and the Ph.D. degree in Electronic and Computer Engineering, Robotics and Telecommunications in 2016, both from the University of Genova, Italy, with a focus on mobile security. He is currently an Associate Professor at DIBRIS, University of Genova, and co-founder and CEO of Talos, a cybersecurity spin-off of the University of Genova. His research interests include Android and mobile security, software security, and the security of machine learning systems.}

\authorbio{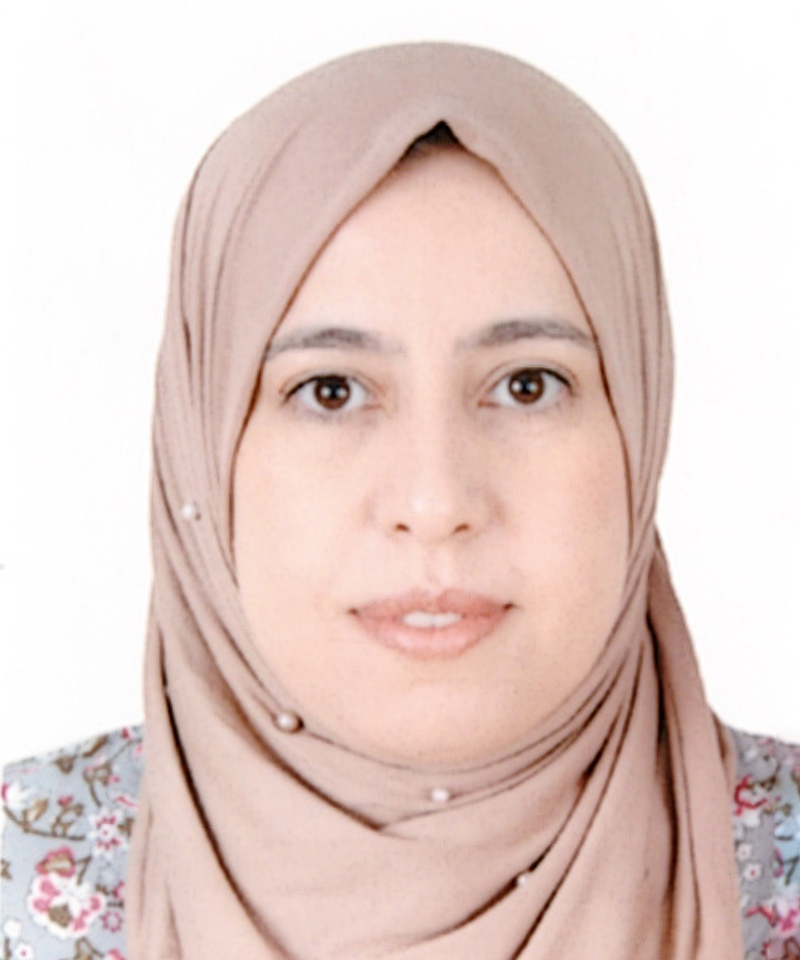}{Meriem Guerar}{received the M.Sc. degree in Information Systems and Networks in 2011 and the Ph.D. degree in Computer Science in 2017, both from the University of Sciences and Technology of Oran, Algeria. She is currently an Assistant Professor at the University of Genova, Italy, and previously held postdoctoral positions at the University of Genova and the University of Padova. Her research interests include mobile and IoT security, privacy, identity management, blockchain, and federated learning.}

\end{document}